\pdfoutput=1
\documentclass[11pt]{article}

\usepackage[final]{acl}

\usepackage{times}
\usepackage{latexsym}

\usepackage[T1]{fontenc}
\usepackage[utf8]{inputenc}

\usepackage{microtype}

\usepackage{inconsolata}

\usepackage{graphicx}

\usepackage{natbib}
\usepackage{graphicx}
\usepackage{subfigure}
\usepackage{wrapfig}
\usepackage{amsmath}
\usepackage{booktabs}
\usepackage[most]{tcolorbox}

\newcommand{\tabincell}[2]{\begin{tabular}{@{}#1@{}}#2\end{tabular}}

\title{Concise and Organized Perception Facilitates Reasoning in Large Language Models}

\author{
  \textbf{Junjie Liu\textsuperscript{1}},
  \textbf{Shaotian Yan\textsuperscript{1}},
  \textbf{Chen Shen\textsuperscript{1}},
  \\
  \textbf{Zhengdong Xiao\textsuperscript{1}},
  \textbf{Liang Xie\textsuperscript{2,1}},
  \textbf{Wenxiao Wang\textsuperscript{3,1}},
  \textbf{Jieping Ye\textsuperscript{1}}
  \\
  \textsuperscript{1}Alibaba Cloud Computing
  \textsuperscript{2}Zhejiang University of Technology
  \textsuperscript{3}Zhejiang University
  \\
  {
    {Corresponding author: Chen Shen (zjushenchen@gmail.com)} %\href{mailto:email@domain}{zjushenchen@gmail.com}
  }
}
\begin{document}
\maketitle
\begin{abstract}
Exploiting large language models (LLMs) to tackle reasoning has garnered growing attention. It still remains highly challenging to achieve satisfactory results in complex logical problems, characterized by plenty of premises within the context and requiring multi-hop reasoning. In particular, the reasoning capabilities of LLMs are brittle to \textit{disorder} and \textit{distractibility}. 
%In this work, from the perspective of information flow, 
In this work, we first examine the mechanism from the perspective of information flow and reveal that LLMs confront difficulties akin to human-like cognitive biases when dealing with disordered and irrelevant content in reasoning tasks. However, in contrast to LLMs, disordered and irrelevant content does not significantly decrease human performance, %which benefits from humans tending to 
as humans have a propensity to distill the most relevant information and systematically organize their thoughts, aiding them in responding to questions.
%to assist them in answering questions. 
%Inspired by the above observation,
Stem from that, we further propose a novel reasoning approach named Concise and Organized Perception (COP). COP carefully analyzes the given statements to identify the most pertinent information while eliminating redundancy efficiently. It then prompts the LLMs in a more organized form that adapts to the model's inference process. By perceiving concise and organized context, the reasoning abilities of LLMs can be better elicited. Extensive experimental results on several popular logical benchmarks (ProofWriter, PrOntoQA, PrOntoQA-OOD, and FOLIO) and mathematical benchmark (DI-GSM) show that COP significantly outperforms previous state-of-the-art methods.
\end{abstract}
%\footnote{Corresponding author}

\section{Introduction}
\label{intro}
%\paragraph{Preprint option}

\begin{figure*}[t]
    \centering
    \includegraphics[width=1.0\linewidth]{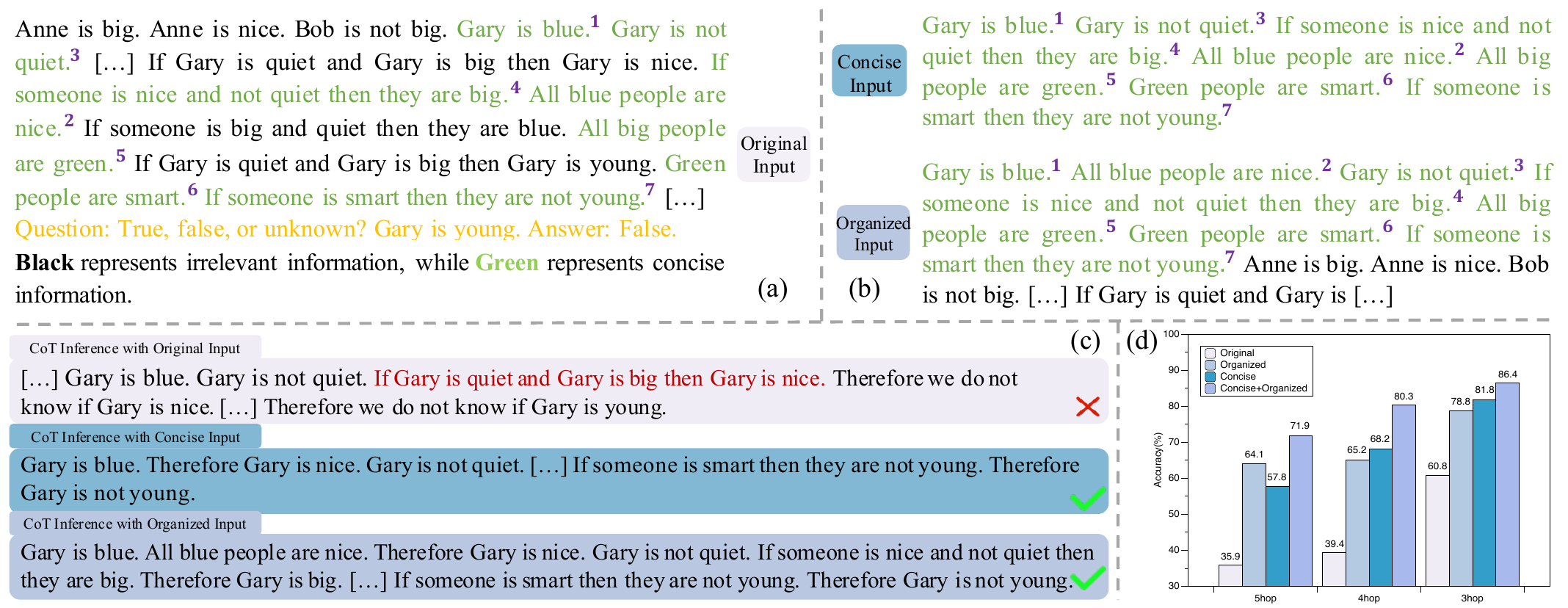}
    \vspace{-10pt}
    \caption{(a) A 5-hop example of ProofWriter dataset, showcasing plenty of premises.% and the question to be answered. 
     Some premises are omitted for brevity. (b) Corresponding reconstruction of concise and organized perception. Superscript serial numbers represent the logical orders according to the gold proof. The concise input contains only relevant information but lacks organizational structure. In contrast, the organized input arranges statements consistently with the gold reasoning path, albeit including redundant information. (c) LLMs outputs. (d) Results of a confirmatory experiment.}
    \label{fig:fig_motivation}
    \vspace{-18pt}
\end{figure*}

The field of large language models (LLMs) has witnessed significant progress in complex reasoning with the advent of Chain-of-thought (CoT) prompting~\citep{wei2022chain} and a series of related works~\citep{qiao2022reasoning,kojima2022large,zhou2023leasttomost,yao2023tree,besta2023graph}. These breakthroughs have yielded remarkable achievements in various applications, including arithmetic, commonsense, symbolic reasoning, etc.~\citep{commonsense,program,evalcode,evalmath,program-aided,generalint}, and have sparked widespread enthusiasm within the community to continuously explore the immense potential of LLMs in tackling complex reasoning tasks. However, the performance of LLMs drastically decreases when handling intricate tasks characterized by plenty of premises within the prompt and requiring multi-hop reasoning. One primary issue is \textit{distractibility}~\citep{irre_context}, where the reasoning capabilities of LLMs are highly susceptible to deterioration when confronted with irrelevant context. Another failure mode that has garnered significant attention recently is \textit{disorder}~\citep{saparov2023language,chen2024premise}, where the performance of LLMs Severely drops when the premise order does not align with the context required in intermediate reasoning steps. 
%Figure~\ref{fig:fig_motivation}(a) presents an example from ProofWriter~\citep{tafjord2021proofwriter}
%, which is a deductive logical dataset 
%(in deductive reasoning, disorder can be further understood as that the premises are not in the same order as the ground truth proof). %The complexity of the context in premises directly affects the difficulty of reasoning. 
Figure~\ref{fig:fig_motivation}(a) shows an example from ProofWriter~\citep{tafjord2021proofwriter}, when the given premises are disordered (in deductive reasoning, disorder can be further understood as that the premises are not in the same order as the ground truth proof) and contain much information irrelevant to the reasoning, the models face a higher risk of selecting the wrong information at some stage. This often leads to an incomplete proof and subsequently an incorrect answer. Figure~\ref{fig:fig_motivation}(c) illustrates such a misleading step, where the original CoT selects the wrong reasoning path (highlighted in red). This observation indicates that LLMs usually struggle with proof planning when irrelevant and disordered content hinders, as also revealed in some contemporaneous works\citep{reversalcurse,irre_context,chen2024premise}.

In this work, we first briefly investigate the underlying mechanism of the influence of disordered and irrelevant content on reasoning. Specifically, we adopt neuron saliency score analysis, which is an important approach for pinpoint the information flow and the crucial interactions between tokens~\citep{dai2021knowledge, hao2021selfattn, labelwords}, and we conclude three phenomena.
%As a important interpretation tool, information flow can pinpoint crucial interactions between tokens. In this work, we further delve deeper into the influence of disordered and irrelevant content on reasoning from the perspective of information flow, 
\textbf{(i)} LLMs struggle to identify the correct entry point of the reasoning path when faced with disorder and distraction. %Intuitively, humans face the same difficulties (i.e., hard to get started) when navigating through a pile of complex information. 
\textbf{(ii)} The current step always highly focuses on the previous step, to the extent that it might even make up non existing premises to accommodate the preceding step. Consequently, it is challenging in allocating sufficient energy to identify the most accurate step. %This phenomenon can be attributed to the model's design, which employs a left-to-right reading paradigm, and closely mimics the natural process of human language comprehension. 
\textbf{(iii)} In addition, salient information flow from irrelevant information renders models prone to distractions, inadvertently causing them to focus on irrelevant content, which ultimately leads to failures in reasoning.

%confronting such complicated circumstances.
%\textbf{(i)} LLMs can identify the correct entry point for reasoning with concise and organized content while failing when faced with disorder and distractibility. Intuitively, such a preference aligns with the human problem-solving processes because locating information within organized and concise sets is considerably more straightforward than navigating through a pile of complex information. \textbf{(ii)} Information flow from the previous step to the current step is salient, which complicates the identification of the next correct reasoning path, and even makes up premises when faced with disordered and irrelevant content. This phenomenon can be attributed to the model's design, which employs a left-to-right reading paradigm, thereby more closely emulating the natural process of human language comprehension. \textbf{(iii)} In addition, salient information flow from irrelevant information makes models distractible, causing models to focus on irrelevant content and leading to reasoning failure.

The information flow analysis reveals how irrelevant and disorganized information can affect model reasoning, and suggests the inertia of LLMs when tackling complex tasks is very similar to the one during human problem-solving process~\citep{humanbiased,failbycog}.
However, in contrast to LLMs, the disordered and irrelevant content does not significantly decline human performance, which benefits from the fact that humans tend to distill the most relevant information and organize their thoughts 
in an orderly manner, such as constructing a mind map, in advance. This allows them to address the question 
%more quickly and 
accurately by referring to the mind map~\citep{ordereffect,orderprefer}. 

Arise from that, we propose a novel reasoning approach named Concise and Organized Perception (COP). Specifically, COP initially performs capturing of locally-related premise segments among the given premises, with the intent to facilitate an initial comprehension of the input context. Next, COP leverages the query question as an anchor to integrate relevant pieces of locally-related premises generated by the first step, creating a tree-like mind map structure that presents global information in an orderly manner and can eliminate irrelevant information. Subsequently, LLMs are prompted by the reconstructed context, which are organized in a progressively ordered manner from the mind map to better adapt to the inference process of the model. 

We believe such reconstruction perceives more concise and organized information, which noticeably reduces the difficulty of LLMs' planning and better elicits the reasoning ability. Figure~\ref{fig:fig_motivation}(b)(c) shows an example where LLMs are empowered to obtain the correct answer. We further conducted a simple confirmatory experiment by randomly selecting 196 samples and reconstructing the context based on the provided ground-truth proofs, as shown in Figure~\ref{fig:fig_motivation}(b)~\footnote{This implementation is merely demonstrative, and differs from the actual method as no ground-truth can be utilized.}. 
The results in Figure~\ref{fig:fig_motivation}(d) demonstrate that combining our approach with the CoT baseline yields a 100\% relative improvement (35.9\% vs 71.9\%) in a 5-hop setting. The results also indicate the complementarity between concise and organized perception. 

We conduct extensive experiments on several popular logical benchmarks (ProofWriter, PrOntoQA, PrOntoQA-OOD), real-world complex logical benchmark (FOLIO) and mathematical benchmark (DI-GSM). The experimental results demonstrate that COP significantly outperforms previous state-of-the-art methods. Specially, COP surpasses the CoT baseline by 9\% on the FOLIO benchmark.

\section{Related work}
\label{related}

LLMs have demonstrated impressive few-shot learning capabilities~\citep{brown2020language, raffel2020exploring, chung2022scaling_bak, ouyang2022training, touvron2023llama}. Recent work has shown that LLMs, combined with in-context learning (ICL) and chain-of-thought (CoT) prompting, are capable of reasoning to an extent~\citep{huang2022towards, qiao2022reasoning, nye2021show, wei2022chain, kojima2022large, lewkowycz2022solving}. However, it still remains highly challenging to achieve satisfactory results in complex logical problems. ~\citep{reallygoodreasoners} explores the logical flaws of LLMs on logical reasoning datasets from four dimensions including answer correctness, explanation correctness, explanation completeness and explanation redundancy. ~\citep{wanABBA} proposes an automatic approach to evaluate the logical reasoning abilities of LLMs based on propositional and predicate logic, which systematically identifies poor logical rules for LLMs' reasoning. In particular, the reasoning capabilities of LLMs are susceptible to deterioration when confronted with inputs that are either arranged in a disordered manner~\citep{saparov2023language,chen2024premise} or peppered with irrelevant information~\citep{irre_context}. Specifically, ~\citep{saparov2023language} investigated how reasoning ability is affected by the traversal direction of the ontology, and ~\citep{chen2024premise} found that the premise order significantly affects LLMs' reasoning performance. In another study, ~\citep{irre_context} observed that the performance of language models tends to decrease when irrelevant context is included in the problem statement. These phenomena aligns with the human preferences for solving logical problems~\citep{ordereffect,orderprefer,mentaldeduction}. Differing from their works, we further investigate the impact of disordered and irrelevant content on reasoning from the perspective of information flow and propose a concise and organized perception approach, drawing inspiration from the perspective of human problem-solving. 

\begin{figure*}[t]
    \begin{center}
    \vspace{-12pt}
    \subfigure[]{
            \includegraphics[width=0.3\linewidth]{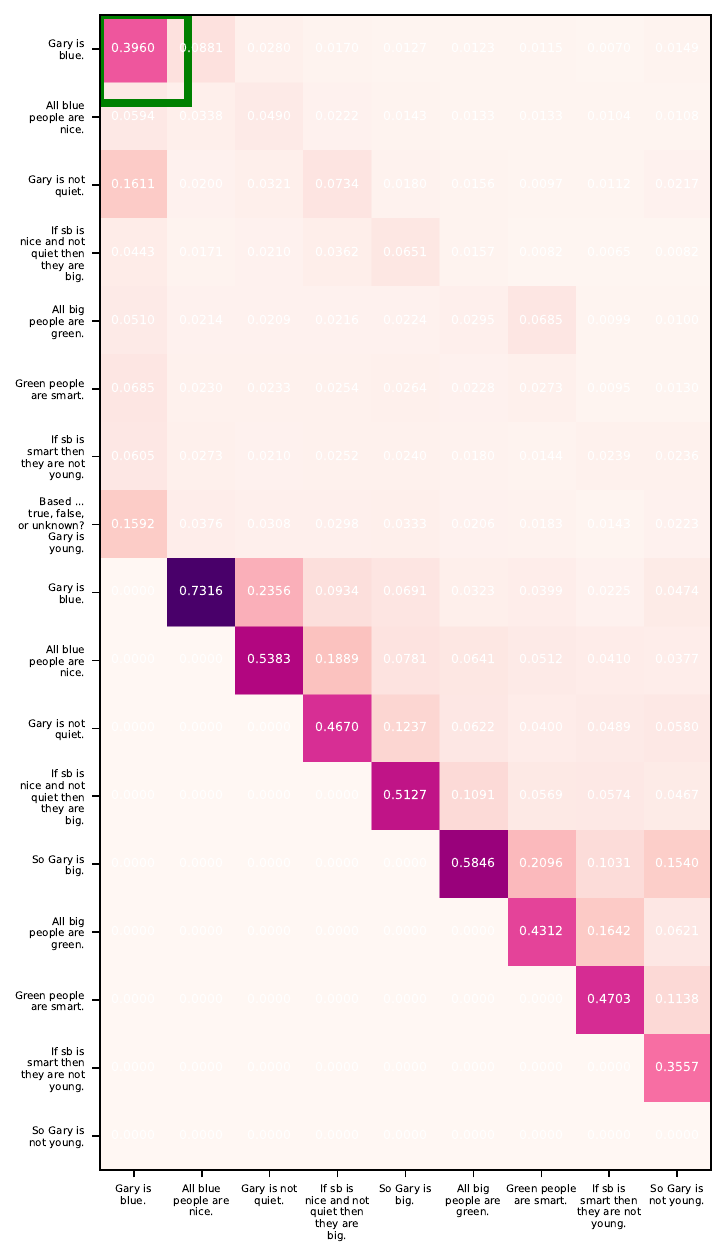} %0.36 % 0.216
    }%
    %\hspace{20pt}
    %\hspace{-10pt}
    \subfigure[]{
            \includegraphics[width=0.3\linewidth]{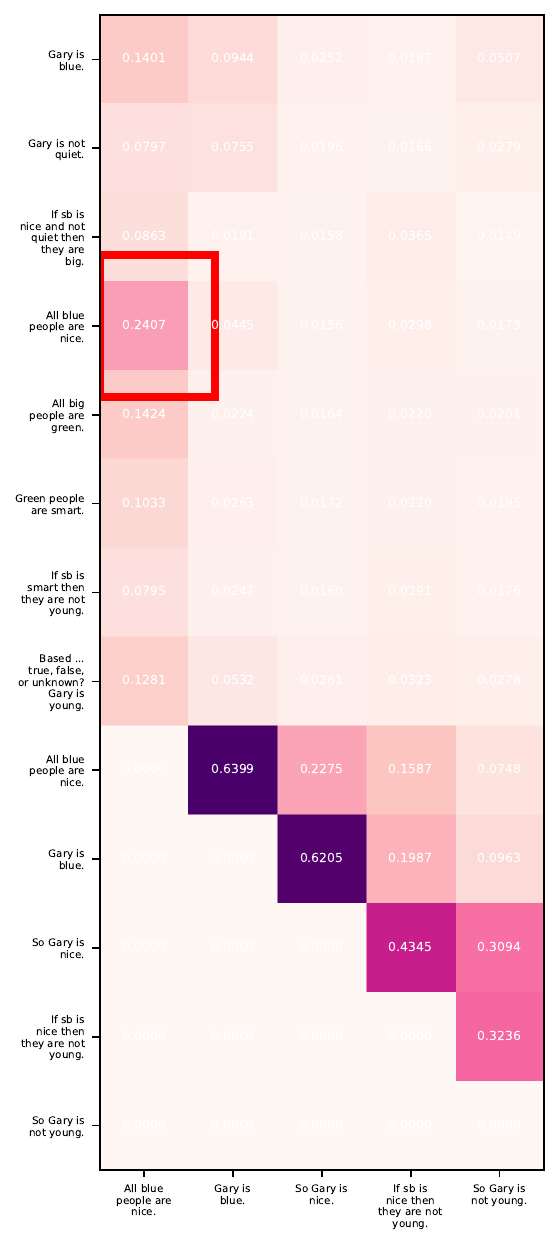} %0.28 %0.168
    }%
    %\hspace{20pt}
    \subfigure[]{
            \includegraphics[width=0.3\linewidth]{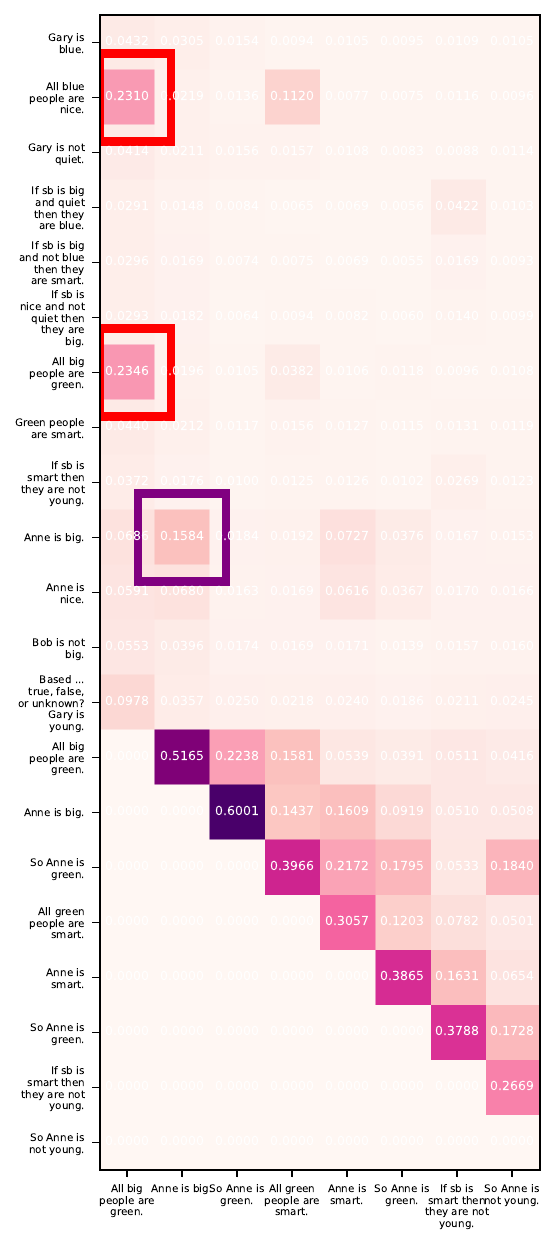} %0.278 % 0.1688
    }%
    %\subfigure[]{
    %        \includegraphics[width=0.35\linewidth]{figs/akk.pdf} %0.278 % 0.35
    %}%
    \vspace{-18pt}
    \end{center}
        \caption{(a)(b)(c) Saliency score analysis on an example of ProofWriter based on shallow layers of Llama-2-13B-Chat. The horizontal coordinate contains the step by step outputs, %the model generates step by step, 
        and the vertical coordinate contains the inputs and outputs. Values in the plot represent saliency scores from column to row, normalized by each column. %(a) A concise and organized example as \textbf{base}. (b) An example with \textbf{disordered} information for comparison. (c) An example with \textbf{irrelevant} information for comparison. 
        }
        %(d) Saliency score analysis on ProofWriter based on shallow and deep layers of Llama-2-13B-Chat. "Base", "Disordered", and "Irrelevant" respectively denote the samples corresponding to the three scenarios depicted in (a)(b)(c). }
        %(d)(1) The saliency scores from the ground-truth reasoning entrance to the first reasoning step. (d)(2) The proportion of samples with the highest saliency score from the ground-truth reasoning entrance to the first reasoning step. (d)(3) The saliency scores from the previous two steps to the current step. (d)(4) The proportion of information flow from relevant and irrelevant information when contains irrelevant information.}
        \label{fig:fig_score_case}
    \vspace{-13pt}
\end{figure*}

\begin{figure}[!h]
    \centering
    \includegraphics[width=1.0\linewidth]{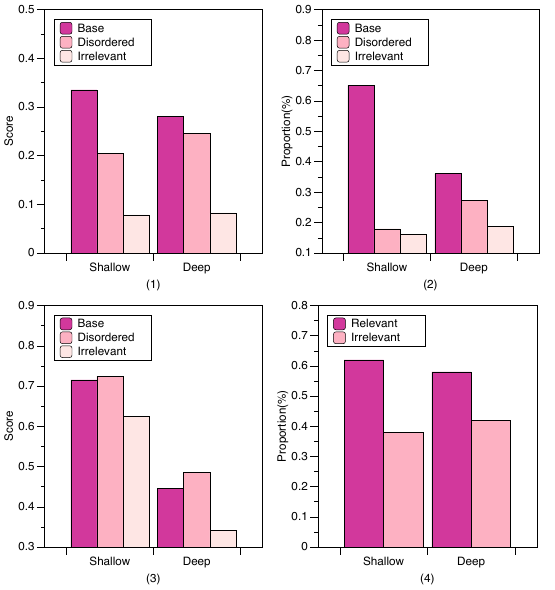}
    \vspace{-10pt}
    \caption{Saliency score analysis on ProofWriter based on shallow and deep layers of Llama-2-13B-Chat. "Base", "Disordered", and "Irrelevant" respectively denote the samples corresponding to the three scenarios depicted in Figure \ref{fig:fig_score_case} (a)(b)(c).}
    \label{fig:fig_score_case_2}
    \vspace{-18pt}
\end{figure}

Benefiting from LLMs' strong logical reasoning ability, some methods~\citep{GSM8K, dalvi2021explaining, zelikman2022star} seek to encourage LLMs to generate reasoning steps explicitly and then produce results in a single stage, while some other methods seek to perform inference at multiple times to complete the tasks~\citep{zhou2023leasttomost,jung2022maieutic}. Several recent works, such as LOGIC-LM~\citep{pan2023logic}, integrate LLMs with symbolic reasoning to improve logical problem-solving. 
%LOGIC-LM first utilizes LLMs to translate a natural language problem into a symbolic formulation. Afterward, a deterministic symbolic solver performs inference on the formulated problem. 
Selection-Inference~\citep{creswell2023selectioninference} alternates between selection and inference to generate a series of casual reasoning steps, and LAMBADA~\citep{kazemi-etal-2023-lambada} develops a backward chaining algorithm to decompose reasoning into sub-modules. In addition to prompting methods, some works aim to fine-tune LLMs to produce the final answer directly, keeping reasoning implicit~\citep{clark2021transformers, lewkowycz2022solving}. In contrast to their works from the perspective of \textit{how to plan} that encourage or teach LLMs how to solve complex logical problems, we introduce the effective and compelling "Concise" and "Organized" strategies to reduce the difficulty of LLMs' reasoning planning and better elicit their reasoning abilities, which can be considered as an alternative angle: reducing the difficulty of planning, or in other words, \textit{easy to plan}. The detailed analysis of 
novelty and effectiveness of COP can be found in Appendix \ref{sec:sec_trynew_appendix}.

\section{Saliency score analysis}
\label{sal_analysis}

As a prevalent paradigm for interpretation, the concept of information flow can be instrumental in dynamically identifying critical interactions amongst tokens~\citep{dai2021knowledge, hao2021selfattn, labelwords}. In this section, we leverage the information flow analysis methodology, predicated upon saliency scores derived from~\citep{labelwords}, to delve deeper into the reasons behind the pronounced degradation in LLMs' performance when confronted with disordered or irrelevant information. We perform analysis on the ProofWriter~\citep{tafjord2021proofwriter} dataset based on Llama-2-13B-Chat~\citep{llama2}, and the detailed saliency score definition and analysis are listed in Appendix~\ref{sec:sec_append_score_analysis}. Figure~\ref{fig:fig_score_case}(a) shows reasoning steps and saliency score analysis on a concise and organized example, which is consistent with the example in Figure~\ref{fig:fig_motivation}. For comparison, the premises of example in Figure~\ref{fig:fig_score_case}(b) are shuffled for analysis the impact of disordered information. Multiple irrelevant premises are added in the example in Figure~\ref{fig:fig_score_case}(c), but the order of the relevant premises is consistent with the example in Figure~\ref{fig:fig_score_case}(a), for analysis the impact of irrelevant information. From Figure~\ref{fig:fig_score_case}, we can observe three phenomena. \textbf{(i)} Model can identify the correct entry point of the reasoning path during the initial step when confronted with concise and organized reasoning content, which is highlighted by the \textcolor{green}{green box} in Figure~\ref{fig:fig_score_case}(a). In contrast, when the model encounters input that is presented in a disordered sequence or contains irrelevant information, it becomes markedly challenging for the model to ascertain an appropriate entry point for reasoning at the initial step, which are highlighted by the \textcolor{red}{red boxes} in Figure~\ref{fig:fig_score_case}(b)(c). \textbf{(ii)} The information flow from the previous step to the current step is salient, as clearly depicted by the diagonal lines in Figure~\ref{fig:fig_score_case}(a)(b)(c). 
This preference for the previous step excessively focuses attention there, complicating the allocation of sufficient effort to identify the subsequent correct step. Furthermore, it may lead to the generation of non-existent premises to cater to the content of the previous step, especially when confronted with disordered and irrelevant content.
In Figure~\ref{fig:fig_score_case}(b), "\textit{if sb is nice then they are not young}" is a fake generated premise, which can be considered as hallucination. \textbf{(iii)} Finally, Figure~\ref{fig:fig_score_case}(c) identifies an adverse effect of irrelevant information on reasoning, with the area highlighted by \textcolor{purple}{purple box} clearly showing a pronounced manifestation of this impact.
This leads to model becoming distractible, causing it to incorporate irrelevant content into the reasoning path, and ultimately resulting in reasoning failure.

\iffalse
\begin{figure}[t]
    \begin{center}
    \subfigure[]{
            \includegraphics[width=0.23\linewidth]{figs/s1.pdf}
    }%
    \subfigure[]{
            \includegraphics[width=0.23\linewidth]{figs/s3.pdf}
    }%
    \subfigure[]{
            \includegraphics[width=0.23\linewidth]{figs/s2.pdf}
    }%
    \subfigure[]{
            \includegraphics[width=0.23\linewidth]{figs/s4.pdf}
    }%
    \end{center}
    \vspace{-10pt}
        \caption{Saliency score analysis on ProofWriter dataset based on shallow and deep layers of Llama-2-13B-Chat. "Base", "Disordered", and "Irrelevant" respectively denote the samples corresponding to the three scenarios depicted in Figure ~\ref{fig:fig_score_case}. (a) The saliency scores from the ground-truth reasoning entrance to the first reasoning step. (b) The proportion of samples with the highest saliency score from the ground-truth reasoning entrance to the first reasoning step. (c) The saliency scores from the previous two steps to the current step. (d) The proportion of information flow from relevant and irrelevant information when contains irrelevant information.}
        \label{fig:fig_score}
\end{figure}
\fi

Further, we conduct quantitative analysis on 1200 samples in the ProofWriter dataset to understand why LLMs' performance significantly decreases when faced with disordered or irrelevant information, from a more holistic viewpoint. Drawing on ~\citep{labelwords}, we conduct a comprehensive analysis of information flow across both the shallow and deep layers. As shown in Figure~\ref{fig:fig_score_case}(d)(1), when inputs are concise and organized, the saliency score from the ground-truth reasoning entrance to the first reasoning step is significantly higher than when faced with disordered or irrelevant content. Figure~\ref{fig:fig_score_case}(d)(2) shows that the proportion of samples with the highest saliency score from the ground-truth reasoning entrance to the first reasoning step is still the highest when inputs are concise and organized. Intuitively, such a preference for the reasoning entry aligns with the problem-solving processes observed in humans because locating information within organized and concise sets is considerably more straightforward than navigating through a pile of complex information. Figure~\ref{fig:fig_score_case}(d)(3) shows the information flow from the previous two steps to the current step, which is also salient as we observed in the diagonal lines in Figure~\ref{fig:fig_score_case}(a)(b)(c). In additional, irrelevant information has dispersed the flow of information as illustrated in Figure~\ref{fig:fig_score_case}(d)(4), which is very similar to the  inertia of humans. However, in contrast to LLMs, the disordered and irrelevant content does not significantly decrease human performance, 
as humans have a propensity to distill the most relevant information and systematically organize their thoughts, aiding them in responding to questions. Thus, draw inspiration from the aspect of human problem-solving, we introduce two effective and compelling strategies ("Concise" and "Organized") to better elicit LLMs' reasoning abilities.

\section{Approach}
\label{approach}

\begin{figure*}[t]
  %\vspace{-13pt}
    \centering
    \includegraphics[width=1.0\linewidth]{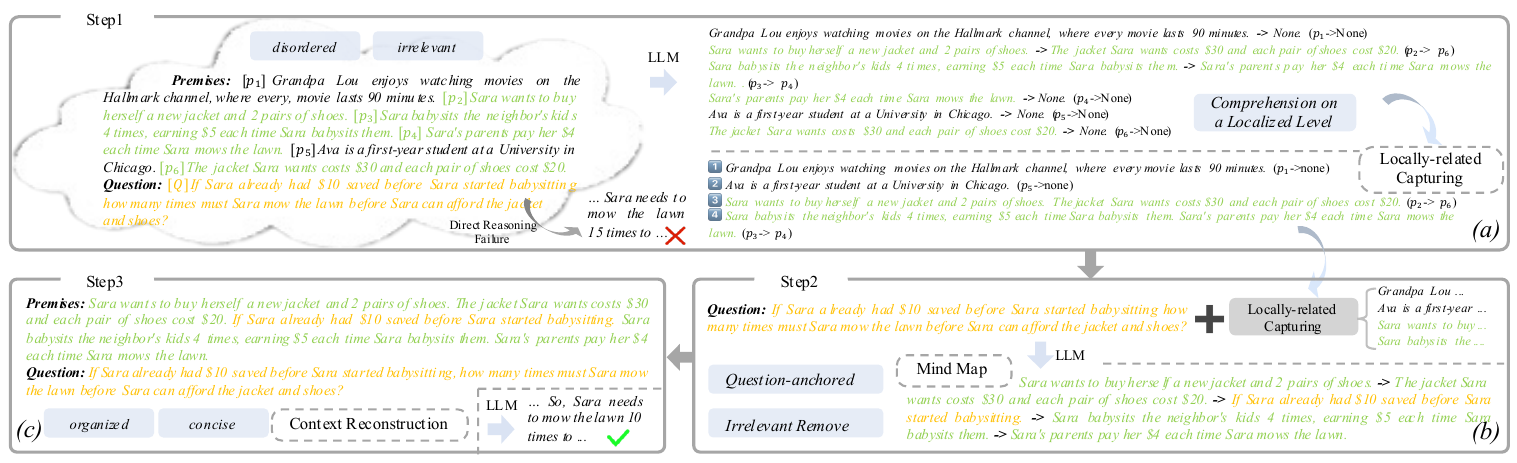}
    %\vspace{-10pt}
    \caption{Overview of the proposed COP with an example on DI-GSM (constructed from GSM8K~\citep{GSM8K}) with disordered and irrelevant premises. Green represents relevant premises [$p_2,p_3,p_4,p_6$], black represents irrelevant premises [$p_1,p_5$], and orange represents the question [$Q$]. Details of DI-GSM are listed in section\ref{sec:sec_datasets}.}\label{fig:fig_framework}
    \vspace{-15pt}
\end{figure*}

We present the Concise and Organized Perception (COP) reasoning approach, aiming to reconstruct concise and organized context as inputting to reduce the difficulty of model reasoning. Given a reasoning context containing plenty of premises $\mathcal{P} = \{p_1, ..., p_n\}$, which may include relevant, irrelevant, or disorganized information, the task is to answer a question $\mathcal{Q}$ based on the given premises.

In order to emulate human capability in processing complex logical reasoning tasks, we propose a three-stage method to effectively tackle disorder and distractibility. Firstly, we seek to capture locally-related premises based on their internal logical or semantic relationships, that is, connecting each single premise between each other to form a series of premise fragments. Such pieces of premise provide an initial structural grasp of the original context on a localized level.
%when faced with intricate problems, humans do not immediately assimilate the information within the given problems. Instead, they tend to scrutinize fragmentary pieces of information, subsequently forging associations among these pieces of information. Building upon this observation, our methodology seeks to construct relationships between the given premises $\mathcal{P}$ in the first step, with the intent to facilitate an initial comprehension of the given information on a localized level. 
Secondly, in the face of several independent pieces, there is a crucial need for holistic systematization to foster comprehension at a global scale. Our approach leverages the question $\mathcal{Q}$ as an anchor to identify relevant fragments and integrate them into a whole tree-like mind map. This structure not only presents global information in an orderly manner but also discards any irrelevant premises.
Subsequently, owing to the progressively organized manner of the mind map, COP creates a more concise and organized reasoning context that can be easily adapted to the inference process of LLMs. The details of these stages will be described in the following subsections.

%As illustrated in Figure~\ref{fig:fig_framework}, the proposed Concise and Organized Perception (COP) initially creates concept maps that highlight the hierarchical relationships among the provided premises to obtain a comprehensive understanding of the problem context. Next, based on the provided question, COP identifies relevant contexts on the concept maps and generates a mind map-like structure based on the question. Subsequently, owing to the progressively organized manner of the mind map, COP creates a more concise and organized reasoning context which can be easily adapted to the inference process of LLMs. The details of these steps will be described in the following subsections.

\subsection{Capturing of locally-related premises}
\label{sec:sec_concept}
% LLM-driven (logical connect or simple connect)
It is generally not a wise strategy to hastily answer questions without fully grasping the given context when performing reasoning tasks; otherwise, it easily leads to inaccurate or incomplete reasoning. Therefore, instead of starting with looking for relevant clues step by step based on each single premise as previous methods (e.g., SI and LAMBADA) do, the first step of the proposed COP is capturing locally-related premises to form a series of premise fragments. 
This allows for an initial structural grasp of the context, facilitating the reconstruction of the context in later steps.

%For example, "\textit{The dog likes the cat}" can be changed into "\textit{dog(like,cat)}", where "\textit{dog}", "\textit{like}" and "\textit{cat}" are the subject, predicate and object respectively. Such uniformly formatted premises can even be connected directly using external tools. 

As mentioned previously, complex reasoning problems involve a set of premises $\mathcal{P}$. Imitating the process of human beings organizing thoughts, capturing of locally-related premises can be effectively achieved by employing directed edges to connect premises that bear relevance to one another. For example, an edge $p_i \rightarrow p_j$ connects premise $p_i$ to premise $p_j$ with the locally-related direction from $i$ to $j$. This approach facilitates a localized understanding of the relationship between the premises $p_i$ and $p_j$. For logical premises, especially modus ponens ($p_a$: if A then B; $p_b$: if B then C), capturing of locally-related premises can be performed by leveraging directed edges to connect each premise to premises whose consequents satisfy one or more of the conditions specified in the current premise. In the given case, premise $p_b$ can be directly connected to premise $p_a$. 
%In addition, for these logical premises with fixed formats, a more direct way is to prompt LLMs to simplify the whole premises into a unified format that is helpful for LLMs to understand the relationship between premises. 
Besides logical premises, capturing of locally-related premises can be performed through semantic correlation, temporal correlation, etc. Figure~\ref{fig:fig_framework}(a) illustrates an instance consisting of premises with semantic correlation. Benefiting from LLMs' powerful in-context learning and semantic understanding ability, our proposed method encourages LLMs to perform capturing locally-related premises by searching relevant premises for each premise in the given context, which can be prompted with few-shot examples. As shown in Figure~\ref{fig:fig_framework}(a), there are six premises in the original problems, which are disordered and contain two irrelevant premises. Through the understanding of semantic correlation, LLMs adeptly identified and correlated relevant premises within vast premises. For example, $p_2$ is connected to $p_6$ while $p_1$ is connected to "None" in Figure~\ref{fig:fig_framework}(a). After capturing locally-related premises for each premise, different pieces of locally-related premises can be integrated again through the connection directions between premises, ultimately forming a initial understanding of the original problem's context. Figure~\ref{fig:fig_framework}(a) shows the four pieces of locally-related premises after integration, in which the relevance between premises is much clearer than that of the original input. %Capturing of locally-related premises can be regarded as a preparatory step similar to the pre-processing that humans undertake before addressing complex tasks. 
The detailed prompts used in this step are listed in the Appendix~\ref{sec:sec_append_prompts}.

\subsection{Generation of mind map}
% LLM-driven (question-based search, logical search or question-based simple depth search)
The above captured locally-related premise fragments
%In section~\ref{sec:sec_concept}, locally-related premise fragments representing the preliminary structural understanding of the given reasoning context are captured. These captured pieces 
are independent and cannot be directly merged. Serving as an anchor point, the query question functions as a connecting bridge. Upon receiving a question, relevant clues can be identified among the pieces of locally-related premises,  allowing for the creation of a tree-like mind map structure relevant to the question. This process effectively eliminates irrelevant information and presents a coherent global understanding of the relevant information in an ordered manner.

Specifically, COP encourages LLMs to find all premises centered around the question $\mathcal{Q}$, which can also be prompted with few-shot examples. The detailed prompts used in this step are listed in the Appendix~\ref{sec:sec_append_prompts}. As shown in Figure~\ref{fig:fig_framework}(b), two relevant pieces of locally-related premises are involved, and another two irrelevant ones are discarded. Although "\textit{If Sara already had \$10 saved before Sara started babysitting}" is a logical description of the question, COP can still correlate it to other pieces of locally-related premises, ultimately forming a holistic and ordered structure. Once we find the relevant pieces of locally-related premises, we perform a $D$-depth searching starting from each of them to avoid reasoning loops, where $D$ is the max reasoning depth. In this way, a tree-like mind map based on the given question is constructed. Generating a tree-like mind map structure based on questions as anchor points is a crucial step in our method that facilitates the development of a structured arrangement centered around the core question while eliminating irrelevant information. Such a strategy provides a foundational guarantee for subsequent concise and organized context reconstruction.
%aligning with human problem-solving preferences. 

\subsection{Context reconstruction}
In the generated mind map, the irrelevant premises to the problem are removed, and the most relevant premises to the problem are retained in order. A straightforward approach to perform context reconstruction is to employ a depth-first traversal technique to comprehensively traverse the entire mind map, thereby obtaining organized inputs, as shown in Figure~\ref{fig:fig_framework}(c). The premise order in the reconstructed context aligns with the progression needed for intermediate reasoning steps, thereby better eliciting the reasoning capabilities of LLMs. Moreover, compared with the original reasoning context, the reconstructed one has the advantage of being concise and dramatically reduces the influence of irrelevant information. 

\begin{figure}[htbp]
    \centering
    \includegraphics[width=1.0\linewidth]{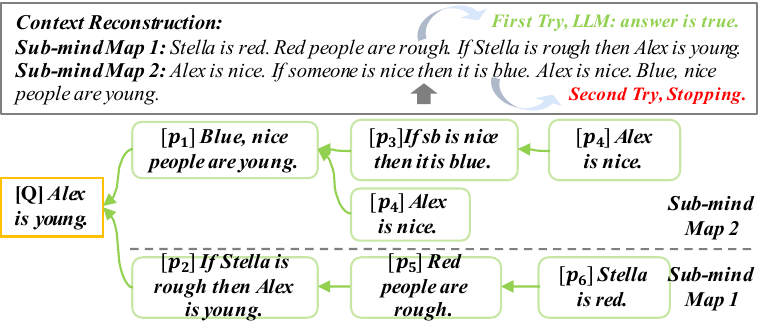}
    %\vspace{-10pt}
    \caption{An example on ProofWriter of sub-mind-map segmentation and context reconstruction.}\label{fig:fig_sub}
    \vspace{-10pt}
\end{figure}

However, a single mind map may contain several sub-mind-maps, especially in scenarios involving logical reasoning. Figure~\ref{fig:fig_sub} presents an example of a mind map consisting of two sub-mind-maps. Given a question "\textit{Alex is young}" and its mind map generated in previous steps, the question is simultaneously relevant with both premise $p_1$ and $p_2$, leading to the natural formation of two sub-mind-maps within the larger mind map. Each sub-mind-map consists of several related premises with directed connections. Therefore, to streamline the reasoning process when faced with multiple sub-mind-maps, it is more efficient to segment the sub-mind-maps before embarking on the final reasoning phase, as opposed to directly engaging with the reconstructed context of the entire mind map.
To answer the given question, a reasoning context for each possible sub-mind-map should be constructed. Subsequently, the reconstructed contexts are successively used to prompt the reasoning of LLMs until a certain answer regarding the given question is made. As illustrated in Figure~\ref{fig:fig_sub}, when the model performs reasoning based on the reconstructed context in sub-mind-map 1 and obtains the exact answer, the reasoning for this question is completed, and the reconstructed context in sub-mind-map 2 is not needed. Such strategy of sub-mind-map segmentation can be regarded as a more refined approach for some exceptional situations.

% Table generated by Excel2LaTeX from sheet 'Sheet14'
\begin{table*}[htbp]
  %\vspace{-10pt}
  %\tabcolsep=1pt
  %\footnotesize
  \tabcolsep=4pt
  \scriptsize
  \centering
  \caption{Comparison of accuracy with various LLMs. "X" means the method has difficulty in running on DI-GSM.}
    \begin{tabular}{c|ccc|ccc|cc|ccc}
    \toprule
    Models & \multicolumn{3}{c|}{GPT-4o} & \multicolumn{3}{c|}{Claude-3-5-Sonnet} & \multicolumn{2}{c|}{Llama-3-70B-Instruct} & \multicolumn{3}{c}{GPT-3.5-Turbo} \\
    \midrule
    Datasets/Methods & FOLIO & DI-GSM & \tabincell{c}{ProofWriter\\-d5} & FOLIO & DI-GSM & \tabincell{c}{ProofWriter\\-d5} & DI-GSM & \tabincell{c}{ProofWriter\\-d5} & \tabincell{c}{ProofWriter\\-d5} & \tabincell{c}{PrOntoQA\\-5hop} & \tabincell{c}{PrOntoQA\\-OOD} \\
    \midrule
    Standard & 58.00  & 80.30  & 61.33  & 64.00  & 79.55  & 65.00  & 76.52  & 55.00  & 41.67  & 49.60  & 43.33  \\
    CoT   & 60.00  & 83.33  & 77.33  & 65.00  & 85.61  & 65.67  & 72.73  & 67.50  & 53.50  & 69.80  & 85.67  \\
    IRRE  & 62.00  & 83.33  & 76.50  & 66.00  & 84.09  & 66.67  & 77.27  & 66.17  & 52.17  & 77.20  & 80.67  \\
    S2A   & 63.00  & 74.24  & 73.33  & 67.00  & 80.30  & 57.33  & 73.48  & 61.00  & 43.00  & 56.00  & 74.67  \\
    SI    & -     & X     & -     & -     & X     & -     & X     & -     & 46.00  & 45.00  & - \\
    LogicLM & 56.00  & X     & 68.33  & 55.00  & X     & 77.83  & X     & 70.17  & 51.40  & 51.58  & - \\
    LAMBABDA & -     & X     & -     & -     & X     & -     & X     & -     & 72.00  & 96.00  & 38.33  \\
    \textbf{COP} & \textbf{65.00 } & \textbf{84.85 } & \textbf{96.50 } & \textbf{68.00 } & \textbf{86.36 } & \textbf{94.50 } & \textbf{78.79 } & \textbf{83.50 } & \textbf{88.67 } & \textbf{99.20 } & \textbf{94.00 } \\
    \bottomrule
    \end{tabular}%
  \label{tab:tab_res}%
\end{table*}%

% Table generated by Excel2LaTeX from sheet 'Sheet15'
\begin{table}[htbp]
  \tabcolsep=4pt
  \scriptsize
  \centering
  \caption{Comparison of accuracy based on Llama-3-8B-Instruct and Qwen2-72B-Chat.}
    \begin{tabular}{c|cc|cc}
    \toprule
    Models & \multicolumn{2}{c|}{Llama-3-8B-Instruct} & \multicolumn{2}{c}{Qwen2-72B-Chat} \\
    \midrule
    Datasets/Methods & DI-GSM & ProofWriter-d5 & DI-GSM & ProofWriter-d5 \\
    \midrule
    Standard & 35.61  & 44.17  & 71.21  & 46.67  \\
    CoT   & 43.18  & 44.33  & 78.79  & 63.83  \\
    IRRE   & 50.00  & 43.17  & 71.97  & 63.17  \\
    S2A & 40.91  & 45.00  & 75.76  & 56.33  \\
    LogicLM & X     & 31.83  & X     & 48.25  \\
    \textbf{COP} & \textbf{52.27 } & \textbf{77.67 } & \textbf{80.30 } & \textbf{75.67 } \\
    \bottomrule
    \end{tabular}%
  \label{tab:tab_res2}%
\end{table}%

% Table generated by Excel2LaTeX from sheet 'Sheet6'
\begin{table}[htbp]
  %\vspace{-5pt}
  \scriptsize%\small
  \centering
  \caption{Comparison of average inference calls and token numbers on FOLIO and ProofWriter.}
    \begin{tabular}{cccc}
    \toprule
    \multicolumn{4}{c}{FOLIO} \\
    \midrule
    Method & Calls & Prompt-tokens & Total-tokens \\
    LogicLM & 4     & 6204.3957 & 7281.5731 \\
    COP   & 3     & 3801.9971 & 4104.9285 \\
    \midrule
    \midrule
    \multicolumn{4}{c}{ProofWriter-5Hop} \\
    \midrule
    Method & Calls & Prompt-tokens & Total-tokens \\
    LAMBADA & 24.71 & 19200.05 & 21922.77 \\
    COP   & 3.87  & 2004.97 & 2440.26 \\
    \bottomrule
    \end{tabular}%
  \label{tab:tab_tokens_merge}%
\end{table}%
\vspace{-5pt}

\section{Experiments}
\label{experiment}
\subsection{Datasets and compared methods}
\label{sec:sec_datasets}
For \textbf{\textit{Real-world Settings}}, we perform experiments on FOLIO and DI-GSM. For \textbf{\textit{Synthetic Settings}}, we perform experiments on ProofWriter, PrOntoQA and PrOntoQA-OOD. Details about datasets can be found in ~\ref{sec:sec_datasets_appendix}. We perform a thorough comparison between our proposed method and the existing state-of-the-art methods (Standard Few-Shot, CoT~\citep{wei2022chain}, IRRE~\citep{irre_context}, S2A~\citep{S2A}, SI~\citep{creswell2023selectioninference}, LogicLM~\citep{pan2023logic} and LAMBADA~\citep{kazemi-etal-2023-lambada}). Details of the compared methods can be found in ~\ref{sec:sec_methods_appendix}.

\subsection{Experimental results}
\subsubsection{Results Analysis}

\textbf{Comparison with SOTA.} Table~\ref{tab:tab_res} and Table~\ref{tab:tab_res2} show the results on different datasets. For SI and LAMBADA, as they just report results based on GPT-3.5-Turbo\citep{kazemi-etal-2023-lambada}, we also conduct experiments on GPT-3.5-Turbo for fair comparison.
%Based on GPT-3.5-Turbo, the results of SI and LAMBADA are taken from \citep{kazemi-etal-2023-lambada}. 
COP consistently achieves the highest accuracy across all experimental settings. Notably, COP outperforms SOTA methods by a large margin on the hardest Depth-5 subset of ProofWriter. It shows a remarkable 65.74\% relative improvement compared to CoT and 23.15\% compared to LAMBADA, which demonstrates the effectiveness of COP. LogicLM converts natural language questions into first-order logical symbolics and performs symbolic reasoning to accomplish logical reasoning tasks. All problems in DI-GSM are mathematical problems, making it difficult for LogicLM to represent them using first-order logic. As shown Table~\ref{tab:tab_res} and Table~\ref{tab:tab_res2}, COP outperforms CoT and LogicLM, while LogicLM is not able to work well on real-world setting DI-GSM (marked as "X"), demonstrating the efficacy of COP. Details of mark "X" are listed in ~\ref{sec:sec_methods_appendix}. The results of LogicLM on FOLIO are reproduced, and the logic program errors in LogicLM are considered as the failure cases in our report. ProofWriter and ProntoQA contain various reasoning depths, detailed results on different reasoning depths are listed in ~\ref{sec:sec_different_part}.  

\textbf{Generalizability.} COP \textit{consistently achieves high accuracy using different LLMs} including GPT-4o, Claude3.5-Sonnet, Llama-3-70B/8B-Instruct, Qwen2-72B-Instruct, crossing  both synthetic setting and real-world setting (FOLIO and DI-GSM). Additional results on more different size open-source and close-source LLMs including Qwen1.5-72B-Chat, Llama-2-13B-Chat, mistral-7b-instruct-v0.3, Gemini-1.0-Pro and GPT-4o-mini can be found in ~\ref{sec:sec_llms}.

\textbf{Efficiency.} Besides, as shown in Table~\ref{tab:tab_tokens_merge}, COP \textit{improves the performance while reducing the cost}. COP utilize two efficient strategies (concise and organized) to identify ordered and related information, reducing the token consumption of irrelevant information and longer reasoning path caused by disorder during reasoning. The detailed comparison of average token usage can be found in ~\ref{sec:sec_append_tokens}.

% We further conduct experiments on DI-GSM to analyze the effectiveness of the components in COP. The results are listed in Table \ref{tab:res_effect}.
\textbf{Ablation study.} We conduct experiments on DI-GSM to analyze the effectiveness of the components in COP. Results are in Table \ref{tab:res_effect}. On the one hand, without the mind map generation step in COP, the accuracy drops (only 43.94), proving the importance of identifying relevant premises around the given question, which is a well-designed step in our COP. One the other hand, we compare with two approaches~\citep{GraphText,Gpt4graph} instead of our mind map generation step. COP is also the best and effective. Details can be found in ~\ref{sec:sec_methods_appendix}.

% Table generated by Excel2LaTeX from sheet 'Sheet8'
\begin{table}[htbp]
  \centering
  \tabcolsep=4pt
  \scriptsize%\small
  %\vspace{10pt}
  \caption{Analysis of the components in COP.}
    \begin{tabular}{cccc}
    \toprule
    \textbf{COP}   & \tabincell{c}{w/o mind map\\ generation} & \tabincell{c}{w/ m1\\~\citep{GraphText}} & \tabincell{c}{w/ m2\\~\citep{Gpt4graph}} \\
    \midrule
    \textbf{53.79} & 43.94 & 41.67 & 43.18 \\
    \bottomrule
    \end{tabular}%
  \label{tab:res_effect}%
\end{table}%

%In addition to GPT-3.5-turbo, due to the computation cost, we select only one hardest synthetic setting ProofWriter-d5 for testing on GPT-4o and Claude3.5-Sonnet, and we select one synthetic setting ProofWriter-d5 and one real-world setting DI-GSM for testing on open-source LLMs. COP \textit{consistently achieves high accuracy using different LLMs} including GPT-4o, Claude3.5-Sonnet, Llama-3-70B/8B-Instruct and Qwen2-72B-Instruct. Additional results on weaker open-source and close-source LLMs including Qwen1.5-72B-Chat, Llama-2-13B-Chat, mistral-7b-instruct-v0.3, Gemini-1.0-Pro and GPT-4o-mini can be found in ~\ref{sec:sec_llms}.

%Unless otherwise specified, all the experimental results of COP are based on GPT-3.5-Turbo~\citep{ouyang2022training}.

%In the next sections, we will analyze COP from several aspects, including concise and organized perception analysis from the perspective of information flow (~\ref{sec:sec_realmetric}), error analysis caused by COP itself (~\ref{sec:sec_error_self}), error analysis when combined with other methods (~\ref{sec:sec_error_other}). In addition, COP consistently achieves high accuracy using different LLMs (including Qwen-Max, Qwen1.5-72B-Chat, Llama-3-70B/8B-Instruct, and Llama-2-13B-Chat), and the detailed results are listed in ~\ref{sec:sec_llms}. 

%with Qwen-Max (qwen-max-0428 version ~\footnote{https://qwenlm.github.io}), Qwen1.5-72B-Chat ~\footnote{https://github.com/QwenLM/Qwen1.5}, Llama-3-70B/8B-Instruct ~\footnote{https://github.com/meta-llama/llama3.git}, and Llama-2-13B-Chat ~\citep{llama2}.

\subsubsection{Failure case analysis}
\label{sec:sec_error_self}
\textbf{COP}. In this section, we focus on the errors caused by COP itself to study its possible flaws. We compare the results of COP on DI-GSM and CoT on DI-GSM with original concise and organized inputs based on GPT-3.5-turbo. The accuracy of COP is 53.79 (71/132), and the accuracy of CoT is 68.94 (91/132). We analyze the differences in the results and find that COP gets 15 more questions correct compared to CoT, and COP fails 35 more questions than CoT. After applying COP, the order of statements in the 15 correct questions is different from the original order, and the changed order generated by COP is more suitable for reasoning, demonstrating the success of organized perception in COP. The error types of the 35 failure cases are shown in Table~\ref{tab:tab_err}. There are 10 cases where the order of the premises caused failure in reasoning. There are 13 cases where the premises are not connected to other premises in the given context. There are 2 cases where the premises are not output in the step of capturing of locally-related premises. These failure cases are attributed to the fact that capturing of locally-related premises is driven by LLMs, and it is difficult for the method to ensure that the generated connections between premises are completely accurate. Similarly, for the remaining 10 cases, some key relevant premises were discarded when generating mind maps, causing failure in reasoning. We present several case studies in ~\ref{sec:sec_case_study} for clearer understanding. We leave the failure caused by these error types as future work. Besides, our approach naturally comes from a different perspective \textit{easy to plan}, it can be seamlessly combined with other popular methods, such as LAMBADA, to further enhance their performance. %Error analysis when combined with LAMBADA can be found in ~\ref{sec:sec_error_other}.

% Table generated by Excel2LaTeX from sheet 'Sheet4'
\begin{table}[htbp]
  \centering
  \scriptsize
  \caption{The detailed failure case analysis on DI-GSM.}
    \begin{tabular}{c|cccc}
    \toprule
    Step & \multicolumn{3}{c}{\tabincell{c}{Capturing of Locally\\-related Premises}} & \tabincell{c}{Generation of\\ Mind Map} \\
    \midrule
    Type  & \tabincell{c}{Connection\\ Order} & \tabincell{c}{Connection\\ Output} & \tabincell{c}{Connection\\ None} & \tabincell{c}{Connection \\Discard} \\
    \midrule
    Errors & 10/35 & 2/35  & 13/35 & 10/35 \\
    \bottomrule
    \end{tabular}%
  \label{tab:tab_err}%
  \vspace{-10pt}
\end{table}% 

\begin{figure}[h]
    \begin{center}
    \subfigure[]{
            \includegraphics[width=0.98\linewidth]{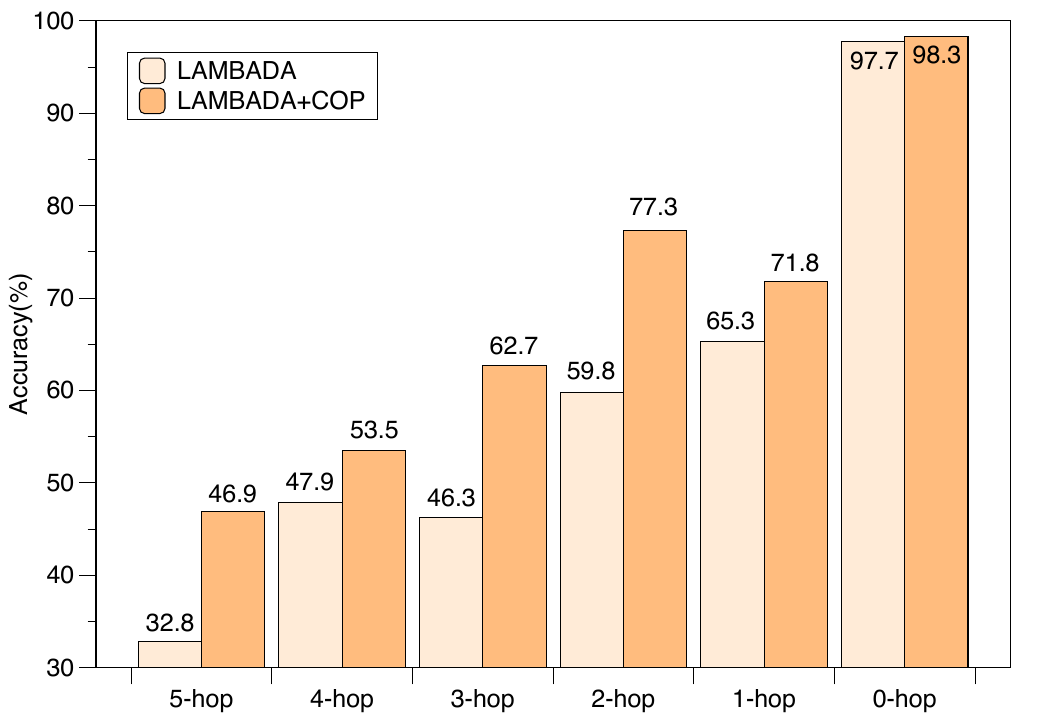}
    }%0.35
    \\
    \subfigure[]{
            \includegraphics[width=0.98\linewidth]{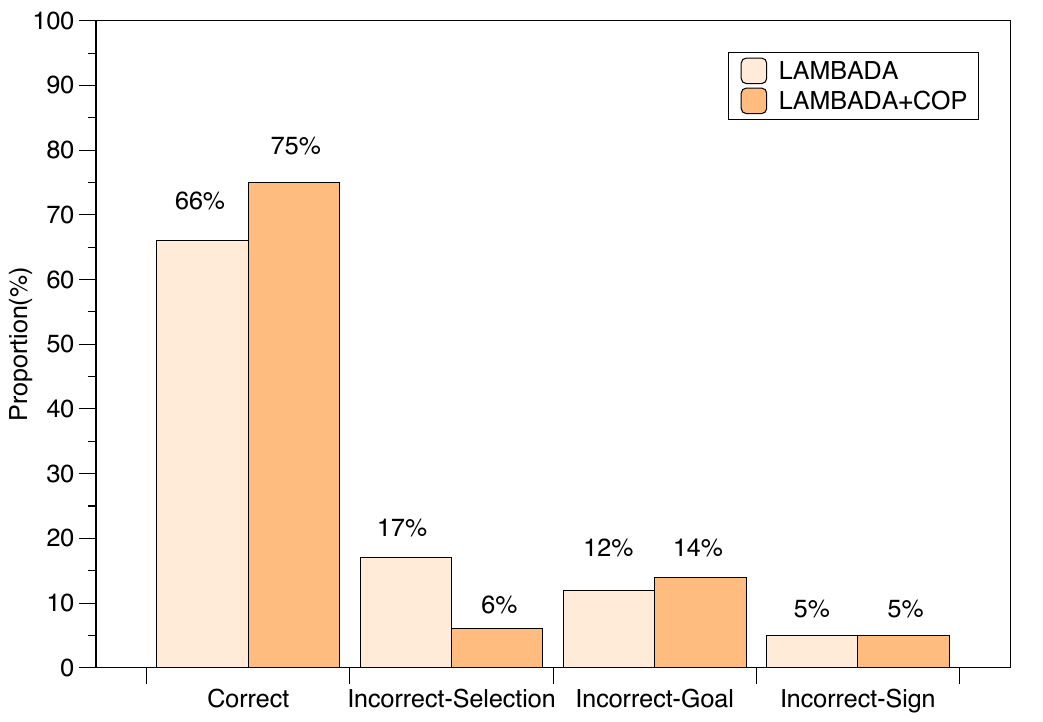}%{figs/err_select.pdf}%{figs/pie1_1.pdf}
    }%0.3
    %\subfigure[]{
    %        \includegraphics[width=0.3\linewidth]{figs/pie2_2.pdf}
    %}%
    \vspace{-20pt}
    \end{center}
        %\caption{(a) Comparison of LAMBADA and LAMBADA+COP on Proofwriter. (b) The proportions of different error reasons of LAMBADA and LAMBADA+COP. There are four steps in LAMBADA, which are Fact Check, Rule Selection, Goal Decomposition and Sign Agreement. "Incorrect-Selection" means that LAMBADA fails in Fact Check or Rule Selection steps. "Incorrect-Goal" means that LAMBADA fails in Goal Decomposition. "Incorrect-Sign" means that LAMBADA fails in Sign Agreement.}
        \caption{(a) Comparison of LAMBADA and LAMBADA+COP on Proofwriter. (b) The proportions of different error reasons of LAMBADA and LAMBADA+COP. There are four steps in LAMBADA, which are Fact Check, Rule Selection, Goal Decomposition and Sign Agreement. "Incorrect-Selection/Goal/  Sign" means that LAMBADA fails in Fact Check or Rule Selection/Goal Decomposition/Sign Agreement steps.}
        \label{fig:fig_append}
    \vspace{-10pt}
\end{figure}

%\subsubsection{Is COP beneficial to other methods ?}
%\label{sec:sec_error_other}
\textbf{COP combined with LAMBADA}. Based on \textit{easy to plan}, our COP can be seamlessly combined with LAMBADA. The performance of LAMBADA and LAMBADA+COP on ProofWriter $d_5$ subset are listed in Figure \ref{fig:fig_append} (a). Compared with the original LAMBADA method, the performance of LAMBADA+COP under different inference depths is improved, proving the effectiveness of COP. In addition, Figure \ref{fig:fig_append}(b) proves that COP can improve the success rate of LAMBADA in the steps that are affected by context redundancy and disorder. Details can be found in ~\ref{sec:sec_error_other}.

%In addition, Figure \ref{fig:fig_append}(b)(c) shows the proportion of correct reasoning and the proportion of different types of incorrect reasoning. We randomly selected 100 test samples from the ProofWriter $d_5$ subset to manually check the error types of incorrect reasoning examples. As shown in the figure, equipped with COP, the proportion of selection errors (including fact check and rule selection modules in LAMBADA) drops significantly. The proportion of goal decomposition errors and sign agreement errors (goal decomposition and sign agreement modules are unaffected by context redundancy and disorder) are almost unchanged, proving that our COP can improve the success rate of other methods in the steps that are affected by context redundancy and disorder.

\section{Conclusion}
\label{conclusion}

%In this study, we investigate the negative impact of disordered and irrelevant content on reasoning from the perspective of information flow , 
% In this work, we first analyze the mechanism from an information flow viewpoint and reveal that LLMs exhibit failure patterns consistent with human cognitive biases when dealing with disordered and irrelevant content in reasoning tasks.
% to effectively tackle disorder and distractibility

In this study, we propose a reasoning approach called Concise and Organized Perception (COP) to handle complex reasoning problems effectively. By combining \textit{Concise} and \textit{Organized} strategies with vanilla CoT, we have achieved state-of-the-art performance on multiple popular reasoning benchmarks. Besides, COP requires significantly fewer inference calls and tokens compared to decomposition-type methods (e.g., LAMBADA), highlighting our superiority in terms of both effectiveness and efficiency. In addition, we investigate the underlying mechanism of the influence of disordered
and irrelevant content on reasoning, and reveal the inherent inertia of LLMs when tackling complex tasks, further supporting the motivation in COP.

\section*{Limitations}
In this work, we focus on two issues including distractibility and disorder. And we focus on the task characterized by plenty of premises, which may contain disordered and irrelevant content. Tasks in which premises do not have an obvious order and do not involve irrelevant information are outside the scope of this paper.

We believe our crucial insight on the proposal of \textit{easy-to-plan} method has broader implications. However, as in error analysis, COP may fails on some examples, but the overall performance improves across different models (results are listed in Main text and Appendix). For more general reasoning tasks, performing robust capturing of locally-related premises and generating more appropriate tree-like mind map structures require further exploration. We plan to address these issues in future research.

%\newpage

%\subsection{References}
\bibliography{sample}

\begin{thebibliography}{48}
\providecommand{\natexlab}[1]{#1}

\bibitem[{Austin et~al.(2021)Austin, Odena, Nye, Bosma, Michalewski, Dohan,
  Jiang, Cai, Terry, Le et~al.}]{program}
Jacob Austin, Augustus Odena, Maxwell Nye, Maarten Bosma, Henryk Michalewski,
  David Dohan, Ellen Jiang, Carrie Cai, Michael Terry, Quoc Le, et~al. 2021.
\newblock Program synthesis with large language models.
\newblock \emph{arXiv preprint arXiv:2108.07732}.

\bibitem[{Berglund et~al.(2023)Berglund, Tong, Kaufmann, Balesni, Stickland,
  Korbak, and Evans}]{reversalcurse}
Lukas Berglund, Meg Tong, Maximilian Kaufmann, Mikita Balesni, Asa~Cooper
  Stickland, Tomasz Korbak, and Owain Evans. 2023.
\newblock The reversal curse: Llms trained on “a is b” fail to learn “b
  is a”.
\newblock In \emph{The Twelfth International Conference on Learning
  Representations}.

\bibitem[{Besta et~al.(2024)Besta, Blach, Kubicek, Gerstenberger, Podstawski,
  Gianinazzi, Gajda, Lehmann, Niewiadomski, Nyczyk et~al.}]{besta2023graph}
Maciej Besta, Nils Blach, Ales Kubicek, Robert Gerstenberger, Michal
  Podstawski, Lukas Gianinazzi, Joanna Gajda, Tomasz Lehmann, Hubert
  Niewiadomski, Piotr Nyczyk, et~al. 2024.
\newblock Graph of thoughts: Solving elaborate problems with large language
  models.
\newblock In \emph{Proceedings of the AAAI Conference on Artificial
  Intelligence}, volume~38, pages 17682--17690.

\bibitem[{Brown et~al.(2020)Brown, Mann, Ryder, Subbiah, Kaplan, Dhariwal,
  Neelakantan, Shyam, Sastry, Askell et~al.}]{brown2020language}
Tom Brown, Benjamin Mann, Nick Ryder, Melanie Subbiah, Jared~D Kaplan, Prafulla
  Dhariwal, Arvind Neelakantan, Pranav Shyam, Girish Sastry, Amanda Askell,
  et~al. 2020.
\newblock Language models are few-shot learners.
\newblock \emph{Advances in neural information processing systems},
  33:1877--1901.

\bibitem[{Bubeck et~al.(2023)Bubeck, Chandrasekaran, Eldan, Gehrke, Horvitz,
  Kamar, Lee, Lee, Li, Lundberg et~al.}]{generalint}
S{\'e}bastien Bubeck, Varun Chandrasekaran, Ronen Eldan, Johannes Gehrke, Eric
  Horvitz, Ece Kamar, Peter Lee, Yin~Tat Lee, Yuanzhi Li, Scott Lundberg,
  et~al. 2023.
\newblock Sparks of artificial general intelligence: Early experiments with
  gpt-4.
\newblock \emph{arXiv preprint arXiv:2303.12712}.

\bibitem[{Chen et~al.(2021)Chen, Tworek, Jun, Yuan, Pinto, Kaplan, Edwards,
  Burda, Joseph, Brockman et~al.}]{evalcode}
Mark Chen, Jerry Tworek, Heewoo Jun, Qiming Yuan, Henrique Ponde de~Oliveira
  Pinto, Jared Kaplan, Harri Edwards, Yuri Burda, Nicholas Joseph, Greg
  Brockman, et~al. 2021.
\newblock Evaluating large language models trained on code.
\newblock \emph{arXiv preprint arXiv:2107.03374}.

\bibitem[{Chen et~al.(2024)Chen, Chi, Wang, and Zhou}]{chen2024premise}
Xinyun Chen, Ryan~Andrew Chi, Xuezhi Wang, and Denny Zhou. 2024.
\newblock Premise order matters in reasoning with large language models.
\newblock In \emph{Forty-first International Conference on Machine Learning}.

\bibitem[{Chung et~al.(2022)Chung, Hou, Longpre, Zoph, Tay, Fedus, Li, Wang,
  Dehghani, Brahma et~al.}]{chung2022scaling_bak}
Hyung~Won Chung, Le~Hou, Shayne Longpre, Barret Zoph, Yi~Tay, William Fedus,
  Eric Li, Xuezhi Wang, Mostafa Dehghani, Siddhartha Brahma, et~al. 2022.
\newblock Scaling instruction-finetuned language models.
\newblock \emph{arXiv preprint arXiv:2210.11416}.

\bibitem[{Clark et~al.(2021)Clark, Tafjord, and
  Richardson}]{clark2021transformers}
Peter Clark, Oyvind Tafjord, and Kyle Richardson. 2021.
\newblock Transformers as soft reasoners over language.
\newblock In \emph{Proceedings of the Twenty-Ninth International Conference on
  International Joint Conferences on Artificial Intelligence}, pages
  3882--3890.

\bibitem[{Cobbe et~al.(2021)Cobbe, Kosaraju, Bavarian, Chen, Jun, Kaiser,
  Plappert, Tworek, Hilton, Nakano et~al.}]{GSM8K}
Karl Cobbe, Vineet Kosaraju, Mohammad Bavarian, Mark Chen, Heewoo Jun, Lukasz
  Kaiser, Matthias Plappert, Jerry Tworek, Jacob Hilton, Reiichiro Nakano,
  et~al. 2021.
\newblock Training verifiers to solve math word problems.
\newblock \emph{arXiv preprint arXiv:2110.14168}.

\bibitem[{Creswell et~al.(2023)Creswell, Shanahan, and
  Higgins}]{creswell2023selectioninference}
Antonia Creswell, Murray Shanahan, and Irina Higgins. 2023.
\newblock Selection-inference: Exploiting large language models for
  interpretable logical reasoning.
\newblock In \emph{The Eleventh International Conference on Learning
  Representations}.

\bibitem[{Dai et~al.(2022)Dai, Dong, Hao, Sui, Chang, and
  Wei}]{dai2021knowledge}
Damai Dai, Li~Dong, Yaru Hao, Zhifang Sui, Baobao Chang, and Furu Wei. 2022.
\newblock Knowledge neurons in pretrained transformers.
\newblock In \emph{Proceedings of the 60th Annual Meeting of the Association
  for Computational Linguistics (Volume 1: Long Papers)}, pages 8493--8502.

\bibitem[{Dalvi et~al.(2021)Dalvi, Jansen, Tafjord, Xie, Smith, Pipatanangkura,
  and Clark}]{dalvi2021explaining}
Bhavana Dalvi, Peter Jansen, Oyvind Tafjord, Zhengnan Xie, Hannah Smith,
  Leighanna Pipatanangkura, and Peter Clark. 2021.
\newblock Explaining answers with entailment trees.
\newblock In \emph{Proceedings of the 2021 Conference on Empirical Methods in
  Natural Language Processing}, pages 7358--7370.

\bibitem[{Dekeyser et~al.(2000)Dekeyser, Schroyens, Schaeken, Spitaels, and
  d’Ydewalle}]{orderprefer}
Mathias Dekeyser, Walter Schroyens, Walter Schaeken, O~Spitaels, and G{\'e}ry
  d’Ydewalle. 2000.
\newblock Preferred premise order in propositional reasoning: Semantic
  informativeness and co-reference.
\newblock \emph{Deductive reasoning and strategies}, pages 73--95.

\bibitem[{Gao et~al.(2023)Gao, Madaan, Zhou, Alon, Liu, Yang, Callan, and
  Neubig}]{program-aided}
Luyu Gao, Aman Madaan, Shuyan Zhou, Uri Alon, Pengfei Liu, Yiming Yang, Jamie
  Callan, and Graham Neubig. 2023.
\newblock Pal: Program-aided language models.
\newblock In \emph{International Conference on Machine Learning}, pages
  10764--10799. PMLR.

\bibitem[{Girotto et~al.(1997)Girotto, Mazzocco, and Tasso}]{ordereffect}
Vittorio Girotto, Alberto Mazzocco, and Alessandra Tasso. 1997.
\newblock The effect of premise order in conditional reasoning: A test of the
  mental model theory.
\newblock \emph{Cognition}, 63(1):1--28.

\bibitem[{Guo et~al.(2023)Guo, Du, Liu, Zhou, He, and Han}]{Gpt4graph}
Jiayan Guo, Lun Du, Hengyu Liu, Mengyu Zhou, Xinyi He, and Shi Han. 2023.
\newblock Gpt4graph: Can large language models understand graph structured
  data? an empirical evaluation and benchmarking.
\newblock \emph{arXiv preprint arXiv:2305.15066}.

\bibitem[{Hagendorff et~al.(2023)Hagendorff, Fabi, and Kosinski}]{humanbiased}
Thilo Hagendorff, Sarah Fabi, and Michal Kosinski. 2023.
\newblock Human-like intuitive behavior and reasoning biases emerged in large
  language models but disappeared in chatgpt.
\newblock \emph{Nature Computational Science}, 3(10):833--838.

\bibitem[{Han et~al.(2022)Han, Schoelkopf, Zhao, Qi, Riddell, Benson, Sun,
  Zubova, Qiao, Burtell, Peng, Fan, Liu, Wong, Sailor, Ni, Nan, Kasai, Yu,
  Zhang, Joty, Fabbri, Kryscinski, Lin, Xiong, and Radev}]{FOLIO}
Simeng Han, Hailey Schoelkopf, Yilun Zhao, Zhenting Qi, Martin Riddell, Luke
  Benson, Lucy Sun, Ekaterina Zubova, Yujie Qiao, Matthew Burtell, David Peng,
  Jonathan Fan, Yixin Liu, Brian Wong, Malcolm Sailor, Ansong Ni, Linyong Nan,
  Jungo Kasai, Tao Yu, Rui Zhang, Shafiq Joty, Alexander~R. Fabbri, Wojciech
  Kryscinski, Xi~Victoria Lin, Caiming Xiong, and Dragomir Radev. 2022.
\newblock Folio: Natural language reasoning with first-order logic.
\newblock \emph{arXiv preprint arXiv:2209.00840}.

\bibitem[{Hao et~al.(2021)Hao, Dong, Wei, and Xu}]{hao2021selfattn}
Yaru Hao, Li~Dong, Furu Wei, and Ke~Xu. 2021.
\newblock Self-attention attribution: Interpreting information interactions
  inside transformer.
\newblock In \emph{Proceedings of the AAAI Conference on Artificial
  Intelligence}, volume~35, pages 12963--12971.

\bibitem[{Hendrycks et~al.(2021)Hendrycks, Burns, Kadavath, Arora, Basart,
  Tang, Song, and Steinhardt}]{evalmath}
Dan Hendrycks, Collin Burns, Saurav Kadavath, Akul Arora, Steven Basart, Eric
  Tang, Dawn Song, and Jacob Steinhardt. 2021.
\newblock Measuring mathematical problem solving with the math dataset.
\newblock In \emph{Thirty-fifth Conference on Neural Information Processing
  Systems Datasets and Benchmarks Track (Round 2)}.

\bibitem[{Huang and Chang(2023)}]{huang2022towards}
Jie Huang and Kevin Chen-Chuan Chang. 2023.
\newblock Towards reasoning in large language models: A survey.
\newblock In \emph{61st Annual Meeting of the Association for Computational
  Linguistics, ACL 2023}, pages 1049--1065. Association for Computational
  Linguistics (ACL).

\bibitem[{Johnson-Laird(2001)}]{mentaldeduction}
Philip~N Johnson-Laird. 2001.
\newblock Mental models and deduction.
\newblock \emph{Trends in cognitive sciences}, 5(10):434--442.

\bibitem[{Jones and Steinhardt(2022)}]{failbycog}
Erik Jones and Jacob Steinhardt. 2022.
\newblock Capturing failures of large language models via human cognitive
  biases.
\newblock \emph{Advances in Neural Information Processing Systems},
  35:11785--11799.

\bibitem[{Jung et~al.(2022)Jung, Qin, Welleck, Brahman, Bhagavatula, Le~Bras,
  and Choi}]{jung2022maieutic}
Jaehun Jung, Lianhui Qin, Sean Welleck, Faeze Brahman, Chandra Bhagavatula,
  Ronan Le~Bras, and Yejin Choi. 2022.
\newblock Maieutic prompting: Logically consistent reasoning with recursive
  explanations.
\newblock In \emph{Proceedings of the 2022 Conference on Empirical Methods in
  Natural Language Processing}, pages 1266--1279.

\bibitem[{Kazemi et~al.(2023)Kazemi, Kim, Bhatia, Xu, and
  Ramachandran}]{kazemi-etal-2023-lambada}
Mehran Kazemi, Najoung Kim, Deepti Bhatia, Xin Xu, and Deepak Ramachandran.
  2023.
\newblock \href {https://doi.org/10.18653/v1/2023.acl-long.361} {{LAMBADA}:
  Backward chaining for automated reasoning in natural language}.
\newblock In \emph{Proceedings of the 61st Annual Meeting of the Association
  for Computational Linguistics (Volume 1: Long Papers)}, pages 6547--6568,
  Toronto, Canada. Association for Computational Linguistics.

\bibitem[{Kojima et~al.(2022)Kojima, Gu, Reid, Matsuo, and
  Iwasawa}]{kojima2022large}
Takeshi Kojima, Shixiang~Shane Gu, Machel Reid, Yutaka Matsuo, and Yusuke
  Iwasawa. 2022.
\newblock Large language models are zero-shot reasoners.
\newblock \emph{Advances in neural information processing systems},
  35:22199--22213.

\bibitem[{Lewkowycz et~al.(2022)Lewkowycz, Andreassen, Dohan, Dyer,
  Michalewski, Ramasesh, Slone, Anil, Schlag, Gutman-Solo
  et~al.}]{lewkowycz2022solving}
Aitor Lewkowycz, Anders Andreassen, David Dohan, Ethan Dyer, Henryk
  Michalewski, Vinay Ramasesh, Ambrose Slone, Cem Anil, Imanol Schlag, Theo
  Gutman-Solo, et~al. 2022.
\newblock Solving quantitative reasoning problems with language models.
\newblock \emph{Advances in Neural Information Processing Systems},
  35:3843--3857.

\bibitem[{Nye et~al.(2021)Nye, Andreassen, Gur-Ari, Michalewski, Austin,
  Bieber, Dohan, Lewkowycz, Bosma, Luan et~al.}]{nye2021show}
Maxwell Nye, Anders~Johan Andreassen, Guy Gur-Ari, Henryk Michalewski, Jacob
  Austin, David Bieber, David Dohan, Aitor Lewkowycz, Maarten Bosma, David
  Luan, et~al. 2021.
\newblock Show your work: Scratchpads for intermediate computation with
  language models.
\newblock In \emph{Deep Learning for Code Workshop}.

\bibitem[{Ouyang et~al.(2022)Ouyang, Wu, Jiang, Almeida, Wainwright, Mishkin,
  Zhang, Agarwal, Slama, Ray et~al.}]{ouyang2022training}
Long Ouyang, Jeffrey Wu, Xu~Jiang, Diogo Almeida, Carroll Wainwright, Pamela
  Mishkin, Chong Zhang, Sandhini Agarwal, Katarina Slama, Alex Ray, et~al.
  2022.
\newblock Training language models to follow instructions with human feedback.
\newblock \emph{Advances in Neural Information Processing Systems},
  35:27730--27744.

\bibitem[{Pan et~al.(2023)Pan, Albalak, Wang, and Wang}]{pan2023logic}
Liangming Pan, Alon Albalak, Xinyi Wang, and William Wang. 2023.
\newblock Logic-lm: Empowering large language models with symbolic solvers for
  faithful logical reasoning.
\newblock In \emph{Findings of the Association for Computational Linguistics:
  EMNLP 2023}, pages 3806--3824.

\bibitem[{Qiao et~al.(2023)Qiao, Ou, Zhang, Chen, Yao, Deng, Tan, Huang, and
  Chen}]{qiao2022reasoning}
Shuofei Qiao, Yixin Ou, Ningyu Zhang, Xiang Chen, Yunzhi Yao, Shumin Deng,
  Chuanqi Tan, Fei Huang, and Huajun Chen. 2023.
\newblock Reasoning with language model prompting: A survey.
\newblock In \emph{Proceedings of the 61st Annual Meeting of the Association
  for Computational Linguistics (Volume 1: Long Papers)}, pages 5368--5393.

\bibitem[{Raffel et~al.(2020)Raffel, Shazeer, Roberts, Lee, Narang, Matena,
  Zhou, Li, and Liu}]{raffel2020exploring}
Colin Raffel, Noam Shazeer, Adam Roberts, Katherine Lee, Sharan Narang, Michael
  Matena, Yanqi Zhou, Wei Li, and Peter~J Liu. 2020.
\newblock Exploring the limits of transfer learning with a unified text-to-text
  transformer.
\newblock \emph{The Journal of Machine Learning Research}, 21(1):5485--5551.

\bibitem[{Saparov and He(2023)}]{saparov2023language}
Abulhair Saparov and He~He. 2023.
\newblock Language models are greedy reasoners: A systematic formal analysis of
  chain-of-thought.
\newblock In \emph{The Eleventh International Conference on Learning
  Representations}.

\bibitem[{Shi et~al.(2023)Shi, Chen, Misra, Scales, Dohan, Chi, Sch\"{a}rli,
  and Zhou}]{irre_context}
Freda Shi, Xinyun Chen, Kanishka Misra, Nathan Scales, David Dohan, Ed~H. Chi,
  Nathanael Sch\"{a}rli, and Denny Zhou. 2023.
\newblock Large language models can be easily distracted by irrelevant context.
\newblock In \emph{Proceedings of the 40th International Conference on Machine
  Learning}, volume 202, pages 31210--31227.

\bibitem[{Tafjord et~al.(2021)Tafjord, Dalvi, and
  Clark}]{tafjord2021proofwriter}
Oyvind Tafjord, Bhavana Dalvi, and Peter Clark. 2021.
\newblock Proofwriter: Generating implications, proofs, and abductive
  statements over natural language.
\newblock In \emph{Findings of the Association for Computational Linguistics:
  ACL-IJCNLP 2021}, pages 3621--3634.

\bibitem[{Talmor et~al.(2019)Talmor, Herzig, Lourie, and Berant}]{commonsense}
Alon Talmor, Jonathan Herzig, Nicholas Lourie, and Jonathan Berant. 2019.
\newblock Commonsenseqa: A question answering challenge targeting commonsense
  knowledge.
\newblock In \emph{Proceedings of the 2019 Conference of the North American
  Chapter of the Association for Computational Linguistics: Human Language
  Technologies, Volume 1 (Long and Short Papers)}, pages 4149--4158.

\bibitem[{Touvron et~al.(2023{\natexlab{a}})Touvron, Lavril, Izacard, Martinet,
  Lachaux, Lacroix, Rozi{\`e}re, Goyal, Hambro, Azhar
  et~al.}]{touvron2023llama}
Hugo Touvron, Thibaut Lavril, Gautier Izacard, Xavier Martinet, Marie-Anne
  Lachaux, Timoth{\'e}e Lacroix, Baptiste Rozi{\`e}re, Naman Goyal, Eric
  Hambro, Faisal Azhar, et~al. 2023{\natexlab{a}}.
\newblock Llama: Open and efficient foundation language models.
\newblock \emph{arXiv preprint arXiv:2302.13971}.

\bibitem[{Touvron et~al.(2023{\natexlab{b}})Touvron, Martin, Stone, Albert,
  Almahairi, Babaei, Bashlykov, Batra, Bhargava, Bhosale et~al.}]{llama2}
Hugo Touvron, Louis Martin, Kevin Stone, Peter Albert, Amjad Almahairi, Yasmine
  Babaei, Nikolay Bashlykov, Soumya Batra, Prajjwal Bhargava, Shruti Bhosale,
  et~al. 2023{\natexlab{b}}.
\newblock Llama 2: Open foundation and fine-tuned chat models.
\newblock \emph{arXiv preprint arXiv:2307.09288}.

\bibitem[{Wan et~al.(2024)Wan, Wang, Yang, Yuan, Huang, He, Jiao, and
  Lyu}]{wanABBA}
Yuxuan Wan, Wenxuan Wang, Yiliu Yang, Youliang Yuan, Jen-tse Huang, Pinjia He,
  Wenxiang Jiao, and Michael~R Lyu. 2024.
\newblock A \& b== b \& a: Triggering logical reasoning failures in large
  language models.
\newblock \emph{arXiv preprint arXiv:2401.00757}.

\bibitem[{Wang et~al.(2023)Wang, Li, Dai, Chen, Zhou, Meng, Zhou, and
  Sun}]{labelwords}
Lean Wang, Lei Li, Damai Dai, Deli Chen, Hao Zhou, Fandong Meng, Jie Zhou, and
  Xu~Sun. 2023.
\newblock Label words are anchors: An information flow perspective for
  understanding in-context learning.
\newblock In \emph{Proceedings of the 2023 Conference on Empirical Methods in
  Natural Language Processing}, pages 9840--9855.

\bibitem[{Wei et~al.(2022)Wei, Wang, Schuurmans, Bosma, Xia, Chi, Le, Zhou
  et~al.}]{wei2022chain}
Jason Wei, Xuezhi Wang, Dale Schuurmans, Maarten Bosma, Fei Xia, Ed~Chi, Quoc~V
  Le, Denny Zhou, et~al. 2022.
\newblock Chain-of-thought prompting elicits reasoning in large language
  models.
\newblock \emph{Advances in neural information processing systems},
  35:24824--24837.

\bibitem[{Weston and Sukhbaatar(2023)}]{S2A}
Jason Weston and Sainbayar Sukhbaatar. 2023.
\newblock System 2 attention (is something you might need too).
\newblock \emph{arXiv preprint arXiv:2311.11829}.

\bibitem[{Xu et~al.(2023)Xu, Lin, Han, Zhao, Liu, and
  Cambria}]{reallygoodreasoners}
Fangzhi Xu, Qika Lin, Jiawei Han, Tianzhe Zhao, Jun Liu, and Erik Cambria.
  2023.
\newblock Are large language models really good logical reasoners? a
  comprehensive evaluation from deductive, inductive and abductive views.
\newblock \emph{arXiv preprint arXiv:2306.09841}.

\bibitem[{Yao et~al.(2024)Yao, Yu, Zhao, Shafran, Griffiths, Cao, and
  Narasimhan}]{yao2023tree}
Shunyu Yao, Dian Yu, Jeffrey Zhao, Izhak Shafran, Tom Griffiths, Yuan Cao, and
  Karthik Narasimhan. 2024.
\newblock Tree of thoughts: Deliberate problem solving with large language
  models.
\newblock \emph{Advances in Neural Information Processing Systems}, 36.

\bibitem[{Zelikman et~al.(2022)Zelikman, Wu, Mu, and
  Goodman}]{zelikman2022star}
Eric Zelikman, Yuhuai Wu, Jesse Mu, and Noah Goodman. 2022.
\newblock Star: Bootstrapping reasoning with reasoning.
\newblock \emph{Advances in Neural Information Processing Systems},
  35:15476--15488.

\bibitem[{Zhao et~al.()Zhao, Zhuo, Shen, Qu, Liu, Bronstein, Zhu, and
  Tang}]{GraphText}
Jianan Zhao, Le~Zhuo, Yikang Shen, Meng Qu, Kai Liu, Michael~M Bronstein,
  Zhaocheng Zhu, and Jian Tang.
\newblock Graphtext: Graph learning in text space.

\bibitem[{Zhou et~al.(2023)Zhou, Sch{\"a}rli, Hou, Wei, Scales, Wang,
  Schuurmans, Cui, Bousquet, Le, and Chi}]{zhou2023leasttomost}
Denny Zhou, Nathanael Sch{\"a}rli, Le~Hou, Jason Wei, Nathan Scales, Xuezhi
  Wang, Dale Schuurmans, Claire Cui, Olivier Bousquet, Quoc~V Le, and Ed~H.
  Chi. 2023.
\newblock Least-to-most prompting enables complex reasoning in large language
  models.
\newblock In \emph{The Eleventh International Conference on Learning
  Representations}.

\end{thebibliography}
%\bibliographystyle{abbrvnat}
%\nocite{Ando2005,andrew2007scalable,rasooli-tetrault-2015}

%%\section*{Acknowledgments}
%% text

%\bibliography{custom}

\appendix

\section{Appendix}
\subsection{Datasets}
\label{sec:sec_datasets_appendix}
\textbf{\textit{Real-world Setting:}} \textbf{(1) FOLIO}~\citep{FOLIO} is a challenging real-world dataset for logical reasoning, written in highly natural wordings and aligned with real-world knowledge. We randomly sampled 100 examples for testing. \textbf{(2) DI-GSM} is a constructed dataset containing \textbf{d}irordered and \textbf{i}rrelevant information, which is one of the latest, complex and representative settings and specially designed for research of distractibility and disorder. Referring to~\citep{chen2024premise}, we first select GSM8K~\citep{GSM8K} test problems with at least five sentences in the problem statements and shuffle the sentences. Besides, we randomly add 2 to 3 irrelevant statements to the questions. The final testing data in DI-GSM contains 132 problems.

\textbf{\textit{Synthetic Setting:}} \textbf{(3) ProofWriter}~\citep{tafjord2021proofwriter} is a commonly used logical reasoning dataset and contains five subsets, named $d5$, $d3$, $d2$, $d1$ and $d0$ respectively. $dx$ part requires $\leq x$ hops for reasoning. We randomly sampled 600 examples in each part. \textbf{(4) PrOntoQA} is a synthetic logical reasoning dataset\footnote{https://github.com/asaparov/prontoqa/tree/v1} and we use three parts $hop5$, $hop3$, and $hop1$ for testing. $hopx$ part requires $x$ hops for reasoning. We randomly sampled 500 examples in each part. \textbf{(5) PrOntoQA-OOD} is another synthetic logical reasoning dataset containing different types of premises. We used the generated data file \textit{generated\_ood\_data.zip} based on open-source code\footnote{https://github.com/asaparov/prontoqa}, and randomly selected 300 samples from the original $hop2$ part for testing. 

\subsection{Compared methods}
\label{sec:sec_methods_appendix}
We perform a thorough comparison between our proposed method and the existing state-of-the-art methods (Standard Few-Shot, CoT~\citep{wei2022chain}, IRRE~\citep{irre_context}, S2A~\citep{S2A}, SI~\citep{creswell2023selectioninference}, LogicLM~\citep{pan2023logic} and LAMBADA~\citep{kazemi-etal-2023-lambada}). As typical methods of teaching model \textit{how to plan}, SI~\citep{creswell2023selectioninference} alternates between selection and inference to generate reasoning steps based on forward chaining, while LAMBADA~\citep{kazemi-etal-2023-lambada} introduces backward chaining for high-level proof planning. LogicLM~\citep{pan2023logic} seek to convert natural language questions into first-order logical symbolics and performs symbolic reasoning to accomplish reasoning tasks. IRRE~\citep{irre_context} explicitly instructs LLMs to ignore irrelevant information in the problem description to perform reasoning. S2A~\citep{S2A} utilizes instruction-tuned LLMs to refine the context by eliminating irrelevant text, allowing for controlled attention focus and deliberate reasoning before generating a response. We set the temperature to 0 for all LLMs and experiments.

\textbf{Details of "X".} In Table~\ref{tab:tab_res} and Table~\ref{tab:tab_res2}, "X" indicates that the method has difficulty in running on DI-GSM. \textbf{(1)} LogicLM is primarily designed for logical reasoning problems. It converts natural language questions into first-order logical symbolics and performs symbolic reasoning to accomplish logical reasoning tasks. In contrast, all problems in DI-GSM are mathematical problems, making it difficult for LogicLM to represent them using first-order logic. Thus, LogicLM cannot work well on DI-GSM. \textbf{(2)} SI and LAMBABDA have difficulty in working on DI-GSM. SI and LAMBABDA are designed for deductive logical reasoning. Here a simple case in deductive logical reasoning, like "If A then B; If B then C;". SI and LAMBABDA perform search based on certain modus ponens (like A, B, C), which is not satisfied on real-world datasets DI-GSM. 

\textbf{Mark "-" in Table~\ref{tab:tab_res}}. Besides, as mentioned in FOLIO, such approaches like SI and LAMBADA, which use superficial strategies and shallow heuristics, can perform well on modus ponens (like ProofWriter), but they perform poorly on FOLIO. The data in FOLIO does not satisfy these easy forms in datasets like ProofWriter. Thus, SI and LAMBADA have difficulty in running on FOLIO. We implemented method LAMBADA and SI according to the details in their paper. The results (based on GPT-4o) in Table \ref{tab:yanzheng} show that LAMBADA and SI do have difficulty in working on FOLIO, and their performance is even worse than CoT. We mark "-" on the FOLIO dataset for SI and LAMBABDA based on other LLMs in our reported results, making the experimental results more rigorous.

% Table generated by Excel2LaTeX from sheet 'Sheet17'
\begin{table}[htbp]
  \small
  \centering
  \caption{Results on FOLIO based on GPT-4o.}
    \begin{tabular}{cccc}
    \toprule
    CoT   & SI    & LAMBABDA & \textbf{COP} \\
    \midrule
    60.00  & 58.00  & 53.00  & \textbf{65.00}  \\
    \bottomrule
    \end{tabular}%
  \label{tab:yanzheng}%
\end{table}%

%SI and LAMBABDA cannot work on both FOLIO and DI-GSM. SI and LAMBABDA are designed for deductive logical reasoning. Here a simple case in deductive logical reasoning, like "If A then B; If B then C;". SI and LAMBABDA perform search based on certain modus ponens (like A, B, C), which is not satisfied on real-world datasets FOLIO and DI-GSM. Thus, SI and LAMBABDA cannot work on both FOLIO and DI-GSM. 

\subsection{Novelty and effectiveness analysis}
\label{sec:sec_trynew_appendix}
\textbf{Novelty}. Distractibility and disorder are two trending issues, which our COP foucus on. It should be noted that IRRE~\citep{irre_context} and S2A~\citep{S2A} seek to only solve distractibility issue in a simple way. And, one of the latest research\citep{chen2024premise} also mentions disorder issue, but it only raises the issue without proposing a solution. In this work, we focus on both these two issues distractibility and disorder. Our paper presents saliency score analysis to demonstrate how disordered and distracting information affects model reasoning for the first time, which provide insights to the communities researching these two issues and supports the motivation in COP. Arise from human problem-solving process and information flow analysis, we propose COP with several key steps including capturing of locally-related premises and mind map generation. And results show that COP can achieve significant performance improvements in complex real-world and multi-hop settings while some other compared methods fail. Information flow analysis and method COP, which is \textbf{novel}, together provide further contributions to distractibility and disorder research. Besides, we compare COP with a direct way based on reordering and filtering information. We conduct an additional experiment based on gpt-3.5-turbo on ProofWriter-D5 and prompt LLM with few-shot examples to directly organize disordered information and ignore irrelevant information in one stage. Results are listed in Table~\ref{tab:res_simple}, and the accuracy based on the direct way is only about 53\%, which is bad and even worse than CoT baseline. On the contrary, the accuracy of COP is significantly improved (88.67\% vs 53.50\%). This benefits from two key steps designed in our COP. So COP is effective and novel. 

% Table generated by Excel2LaTeX from sheet 'Sheet8'
\begin{table}[htbp]
  \centering
  \small
  \caption{The performance comparisons on Proofwriter-d5 between the direct way, CoT and COP.}
    \begin{tabular}{ccc}
    \toprule
    CoT   & Direct-way & \textbf{COP} \\
    \midrule
    53.50  & 53.17 & \textbf{88.67} \\
    \bottomrule
    \end{tabular}%
  \label{tab:res_simple}%
\end{table}%

\textbf{Effectiveness.} We further conduct experiments on DI-GSM to analyze the effectiveness of the components in COP. The results are listed in Table \ref{tab:res_effect2}. On the one hand, without the mind map generation step in COP, the accuracy drops (only 43.94), proving the importance of identifying relevant premises around the given question, which is a well-designed step in our COP. One the other hand, we compare with two approaches~\citep{GraphText,Gpt4graph}. We change graph information into text format to prompt LLM reasoning in \citep{GraphText}(m1). We use Graph Modelling Language(GML) format to prompt LLM reasoning in \citep{Gpt4graph}(m2). Model fails to understand the relations between premises based on graph information in ~\citep{GraphText,Gpt4graph}, which prevents it from identifying the most relevant premises to the problem in order. Overall, the steps designed in COP, which are \textbf{effective}, ensure the performance.

% Table generated by Excel2LaTeX from sheet 'Sheet8'
\begin{table}[htbp]
  \centering
  \tabcolsep=4pt
  \small
  \caption{Effectiveness analysis of the components in COP.}
    \begin{tabular}{cccc}
    \toprule
    \textbf{COP}   & \tabincell{c}{w/o mind map\\ generation} & \tabincell{c}{w/ m1\\~\citep{GraphText}} & \tabincell{c}{w/ m2\\~\citep{Gpt4graph}} \\
    \midrule
    \textbf{53.79} & 43.94 & 41.67 & 43.18 \\
    \bottomrule
    \end{tabular}%
  \label{tab:res_effect2}%
\end{table}%

\subsection{Additional saliency score analysis}
\label{sec:sec_append_score_analysis}

\subsubsection{Saliency score definition}
According to~\citep{labelwords}, saliency score for each element of the attention matrix is defined as:
\begin{equation}\label{equ:score}
\small
    I_l = \sum_{h} {|A_{h,l}^{T}\frac {\partial L(x)} {\partial {A_{h,l}}}|}
\end{equation}
where $A_{h,l}$ is the attention matrix of the $h$-th head in the $l$-th layer, $x$ is the inputs, $L(x)$ is the cross-entropy loss function, saliency score $I_l(i,j)$ is the significance of the information flow from token $i$ to $j$ in the $x$. The information flow from sentence $A$ to sentence $B$ is defined as:
\begin{equation}\label{equ:score_aga}
\small
    IS(A,B) = \frac {1} {|\{lers\}|} \sum_{l \in \{lers\}} (\frac {1} {|t_{a}| |t_{b}|} \sum_{a \in t_{a}} \sum_{b \in t_{b}} I_l(a,b))
\end{equation}
where $t_{a}$ and $t_{b}$ are the token sets of sentences $A$ and $B$ respectively, $|t_{a}|$ and $|t_{b}|$ are the length of $t_{a}$ and $t_{b}$, $\{lers\}$ is the set of candidate layers in LLM. $IS(A,B)$ is normalized according to B in each generation step. Drawing on~\citep{labelwords}, we analyze information flow from the shallow and deep layers of the LLM. Specifically, given a model with $L$ layers, we select first five layers $\{lers\}=\{1,2,3,4,5\}$ for shallow analysis, and select last five layers $\{lers\}=\{L-4,L-3,L-2,L-1,L\}$ for deep analysis. In our experiments, we analyze information flow from multiple aspects:

\begin{enumerate}

      \item The saliency score from the ground-truth reasoning entrance to the first reasoning step is defined as:
\begin{equation}\label{equ:score_p1}
\small
      A_1 = \frac {1} {N} \sum_{i=1}^{N} IS(s_{entrance}^{(i)},g_{1}^{(i)})
\end{equation}
where $s_{entrance}^{(i)}$ is the ground-truth reasoning entrance sentence in inputs of sample $i$, $g_{1}^{(i)}$ is the first generated sentence (outputed by LLMs) of sample $i$, $N$ is the total number of test samples. A larger $A_1$ means that the model can better find the reasoning entrance.
      
      \item The proportion of samples with the highest saliency score from the ground-truth reasoning entrance to the first reasoning step is defined as :
\begin{equation}\label{equ:score_p2}
\small
\begin{split}
      A_2 = \frac {1} {N} \sum_{i=1}^{N} \delta (s_{entrance}^{(i)}, s_{opt}^{(i)}), \\
      s_{opt}^{(i)} = \mathop{\arg\max}\limits_{s_j^{(i)}} IS(s_j^{(i)},g_{1}^{(i)})
\end{split}
\end{equation}
where $s_{entrance}^{(i)}$ is the ground-truth reasoning entrance sentence in inputs of sample $i$, $s_j^{(i)}$ is the $j$-th sentence in inputs of sample $i$, $g_{1}^{(i)}$ is the first generated sentence (outputed by LLMs) of sample $i$, $\delta$ is the Kronecker function, $N$ is the total number of test samples. A larger $A_2$ means that the model can also better find the reasoning entrance.

      \item The saliency score from the previous two steps to the current step is defined as:
\begin{equation}\label{equ:score_p3}
\small
      A_3 = \frac {1} {NK} \sum_{i=1}^{N} \sum_{k=1}^{K} IS(g_{k-2}^{(i)},g_{k}^{(i)}) + IS(g_{k-1}^{(i)},g_{k}^{(i)})
\end{equation}
where $g_{k}^{(i)}$ is the $k$-th generated sentence (outputed by LLMs) of sample $i$, $K$ is the total number of generated sentences, $N$ is the total number of test samples. The value of $A_3$ indicates the information from the previous two steps to the current step. A larger $A_3$ means the model is more likely to pay attention to the information in previous two steps.

      \item The proportion of information flow from relevant and irrelevant information when contains irrelevant information is defined as:
\begin{equation}\label{equ:score_p4}
\small
\begin{split}
      &A_4 = \frac {r} {r+1}, \\
      &r = {\frac{1}{N} \sum_{i=1}^{N} {\frac {\frac {1} {K|j_{irre}^{(i)}|} \sum_{k=1}^{K} \sum_{j \in j_{irre}^{(i)}} IS(s_j^{(i)},g_{k}^{(i)})} {\frac {1} {K|j_{re}^{(i)}|} \sum_{k=1}^{K} \sum_{j \in j_{re}^{(i)}} IS(s_j^{(i)},g_{k}^{(i)})} }} \\
\end{split}
\end{equation}
where $g_{k}^{(i)}$ is the $k$-th generated sentence (outputed by LLMs) of sample $i$, $K$ is the total number of generated sentences, $j_{re}^{(i)}$ is the index sets of relevant ground-truth reasoning sentences in inputs of sample $i$, $j_{irre}^{(i)}$ is the index sets of irrelevant sentences in inputs of sample $i$, $|j_{re}^{(i)}|$ and $|j_{irre}^{(i)}|$ are the length of $j_{re}^{(i)}$ and $j_{irre}^{(i)}$, $N$ is the total number of test samples. Then, the saliency score from relevant information is defined as $1-A_4$. A larger $A_4$ means that the information flow from irrelevant information is more salient, and the model is more likely to foucs on irrelevant information.
  
\end{enumerate}

\begin{figure*}[htbp]
    \begin{center}
    \subfigure[]{
            \includegraphics[width=0.23\linewidth]{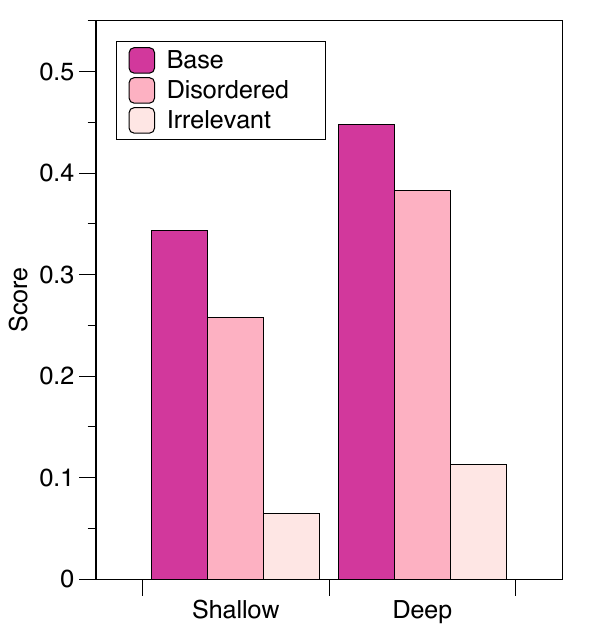}
    }%
    \subfigure[]{
            \includegraphics[width=0.23\linewidth]{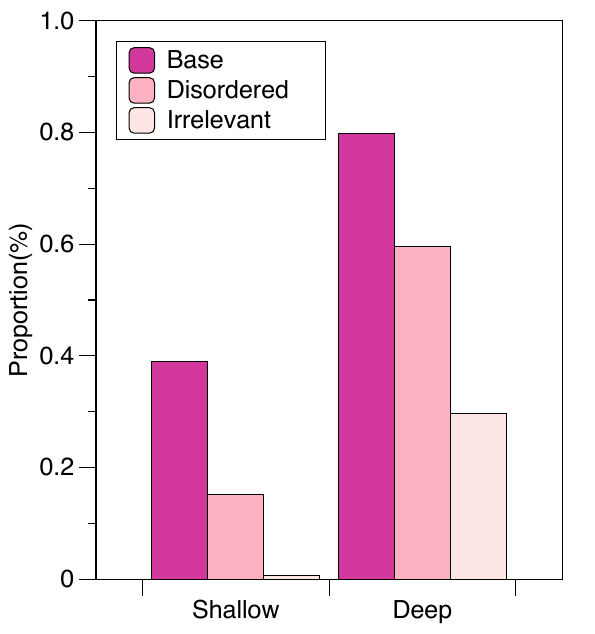}
    }%
    \subfigure[]{
            \includegraphics[width=0.23\linewidth]{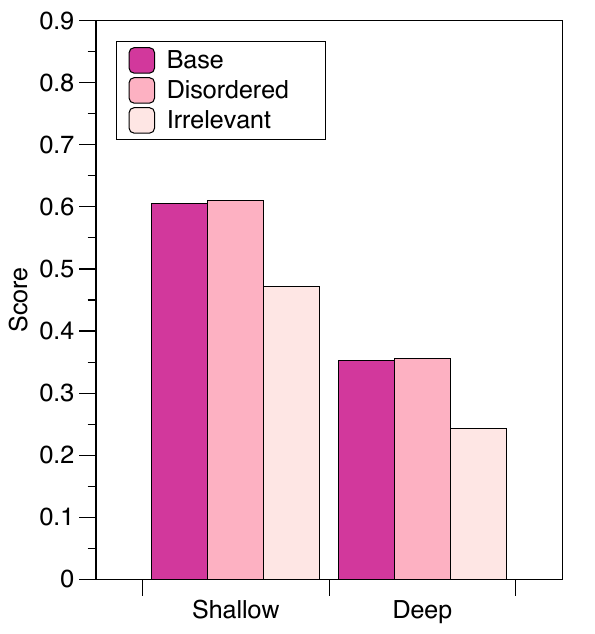}
    }%
    \subfigure[]{
            \includegraphics[width=0.23\linewidth]{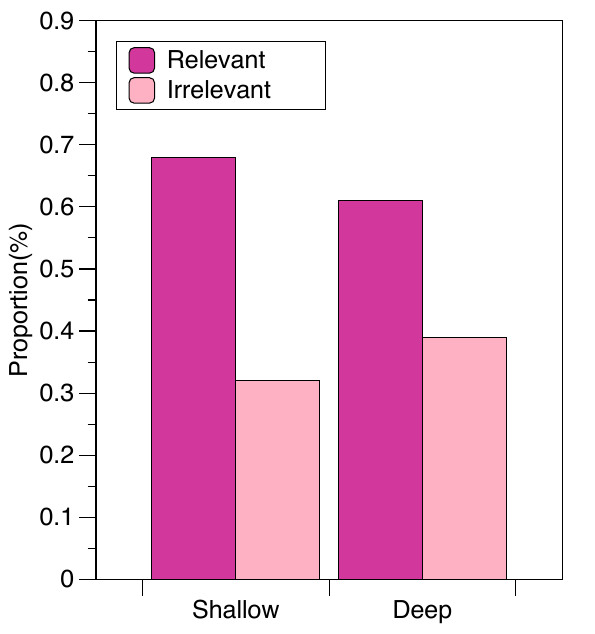}
    }%
    \end{center}
    \vspace{-10pt}
        \caption{Saliency score analysis on ProofWriter based on Qwen-14B-Chat. (a) The saliency scores from the ground-truth reasoning entrance to the first reasoning step ($A_1$). (b) The proportion of samples with the highest saliency score from the ground-truth reasoning entrance to the first reasoning step ($A_2$). (c) The saliency scores from the previous two steps to the current step ($A_3$). (d) The proportion of information flow from relevant and irrelevant information when contains irrelevant information ($A_4$).}
        \label{fig:fig_score_qwen}
\end{figure*}

\begin{figure*}[htbp]
    \begin{center}
    \subfigure[]{
            \includegraphics[width=0.23\linewidth]{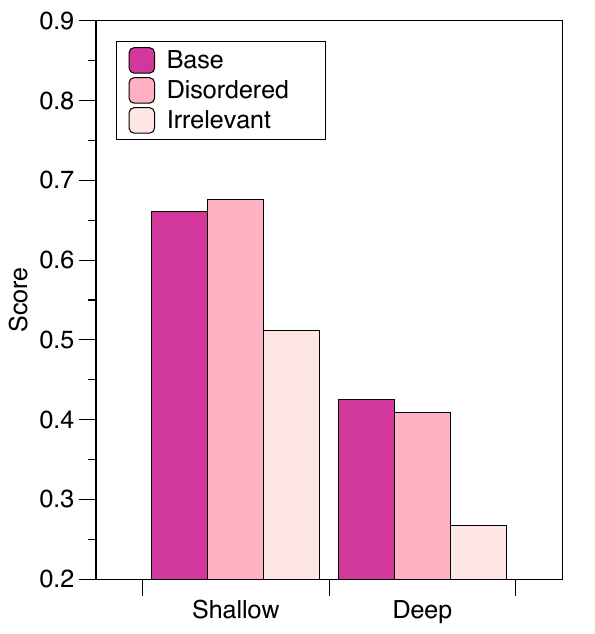}
    }%
    \subfigure[]{
            \includegraphics[width=0.23\linewidth]{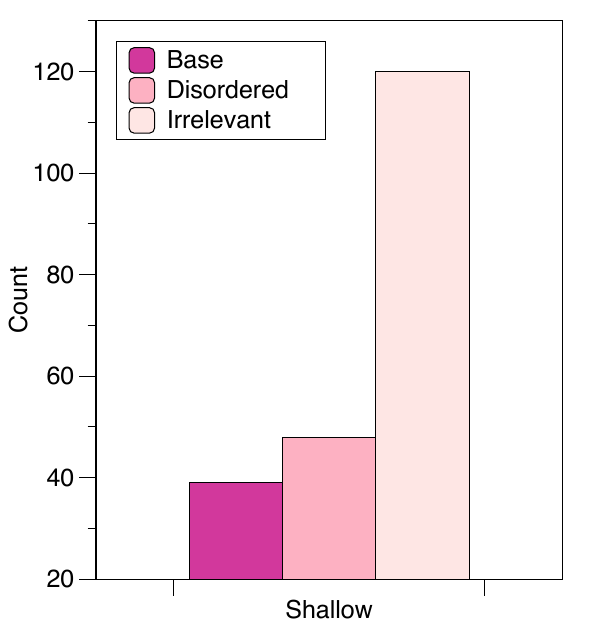}
    }%
    \subfigure[]{
            \includegraphics[width=0.23\linewidth]{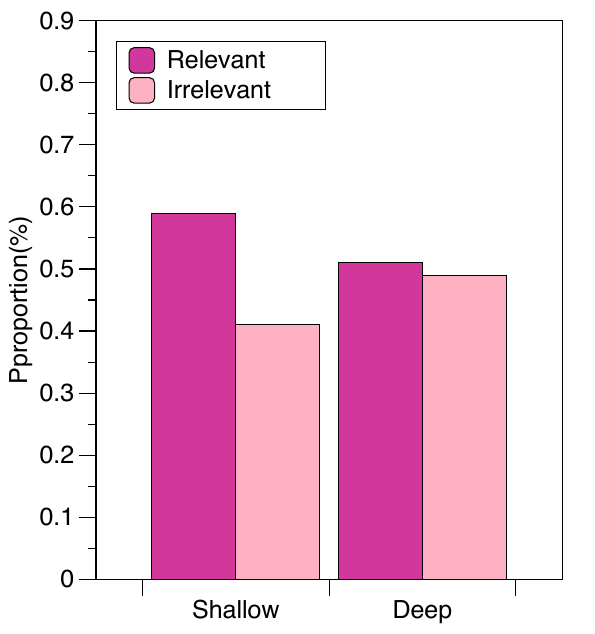}
    }%
    \end{center}
    \vspace{-10pt}
        \caption{Generated fake premises analysis on ProofWriter based on Qwen-14B-Chat. (a) The saliency scores from the previous two steps to the current step ($A_3$) when generating a fake premise that is not in the given premises. (b) The number of the generated fake premises. (c) The proportion of information flow from relevant and irrelevant information when contains irrelevant information ($A_4$) when generating a fake premise that is not in the given premises.}
        \label{fig:fig_gen_qwen_count}
\end{figure*}

\begin{figure*}[htbp]
    \begin{center}
    \subfigure[]{
            \includegraphics[width=0.23\linewidth]{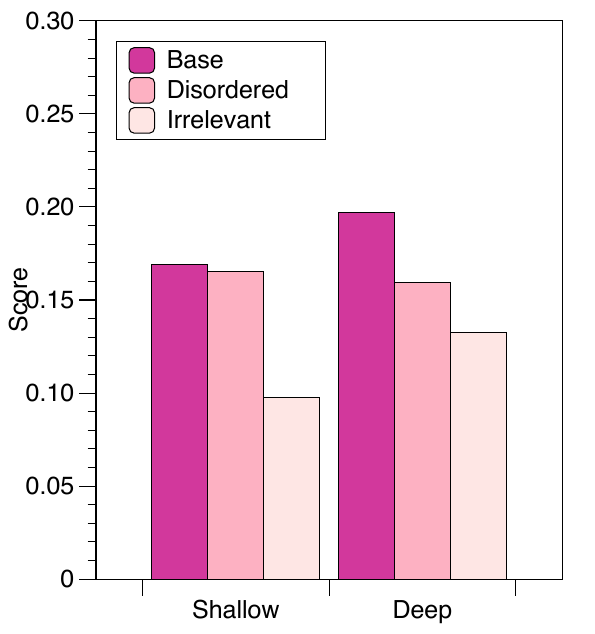}
    }%
    \subfigure[]{
            \includegraphics[width=0.23\linewidth]{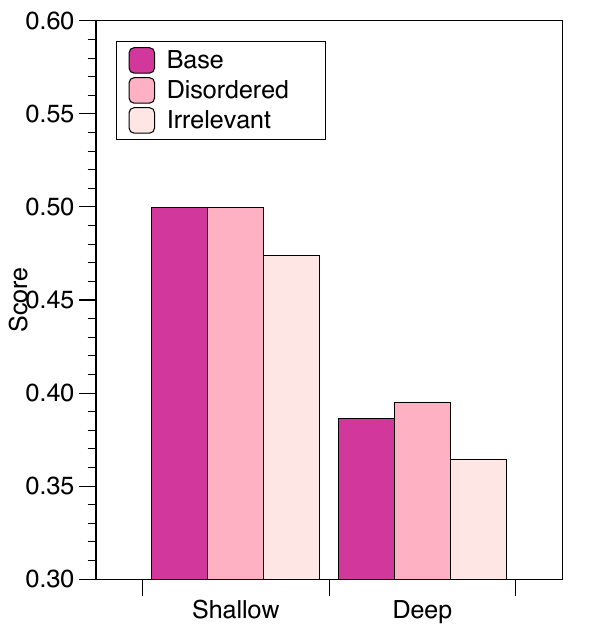}
    }%
    \subfigure[]{
            \includegraphics[width=0.23\linewidth]{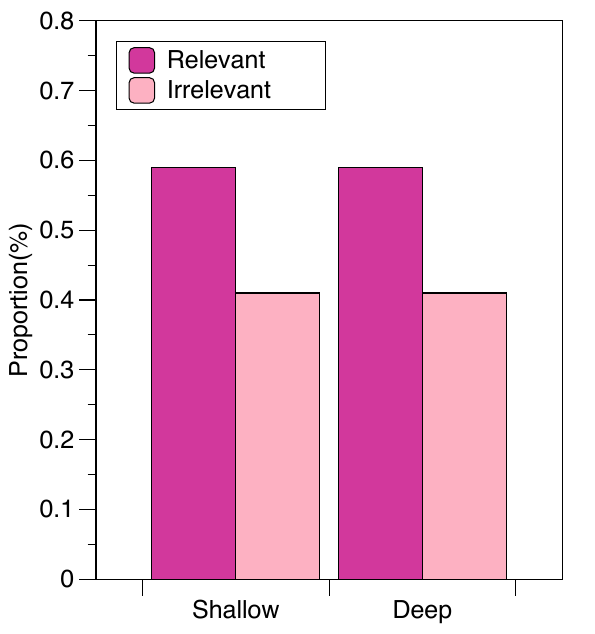}
    }%
    \end{center}
    \vspace{-10pt}
        \caption{Saliency score analysis on DI-GSM based on Llama-2-13B-Chat. (a) The saliency scores from the ground-truth reasoning entrance to the first reasoning step ($A_1$). (b) The saliency scores from the previous two steps to the current step ($A_3$). (c) The proportion of information flow from relevant and irrelevant information when contains irrelevant information ($A_4$).}
        \label{fig:fig_score_llama_gs}
\end{figure*}

\begin{figure*}[htbp]
    \begin{center}
    \subfigure[]{
            \includegraphics[width=0.23\linewidth]{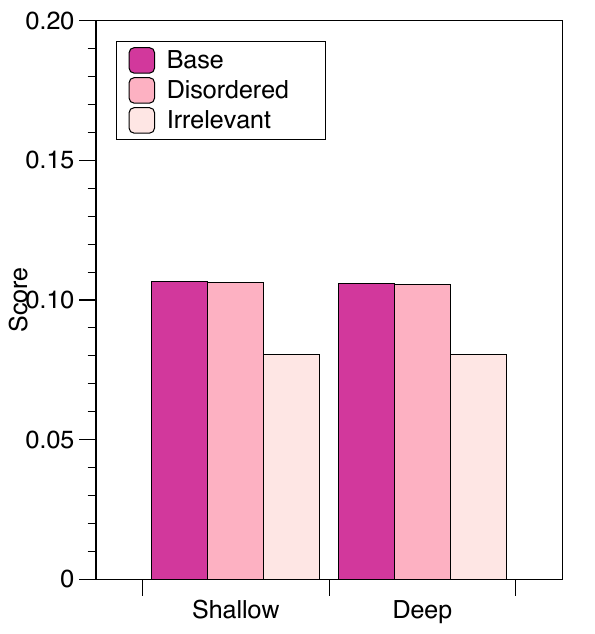}
    }%
    \subfigure[]{
            \includegraphics[width=0.23\linewidth]{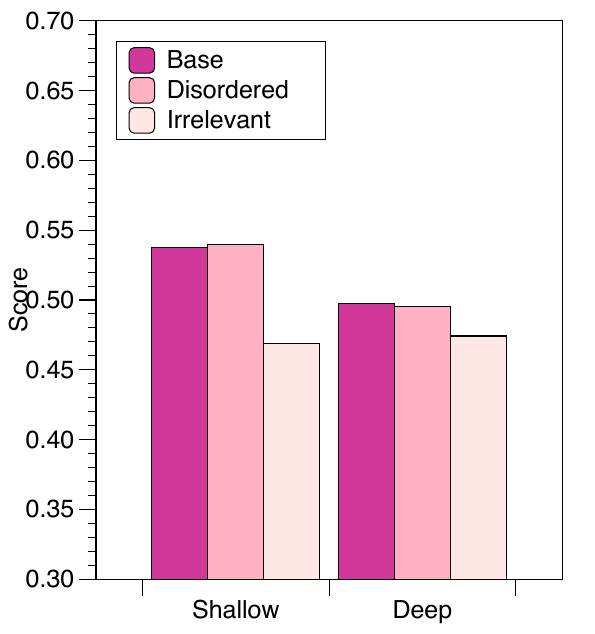}
    }%
    \subfigure[]{
            \includegraphics[width=0.23\linewidth]{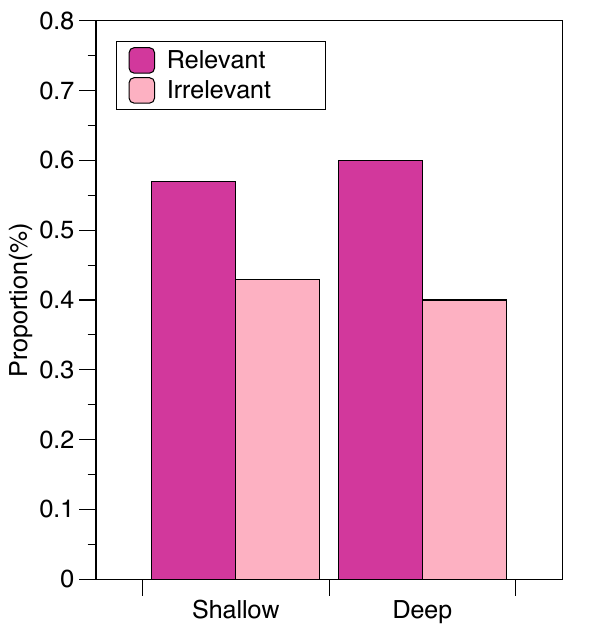}
    }%
    \end{center}
    \vspace{-10pt}
        \caption{Saliency score analysis on DI-GSM based on Qwen-14B-Chat. (a) The saliency scores from the ground-truth reasoning entrance to the first reasoning step ($A_1$). (b) The saliency scores from the previous two steps to the current step ($A_3$). (c) The proportion of information flow from relevant and irrelevant information when contains irrelevant information ($A_4$).}
        \label{fig:fig_score_qwen_gs}
\end{figure*}

\begin{figure*}[htbp]
  \vspace{-50pt}
  \centering
  \includegraphics[width=0.8\textwidth]{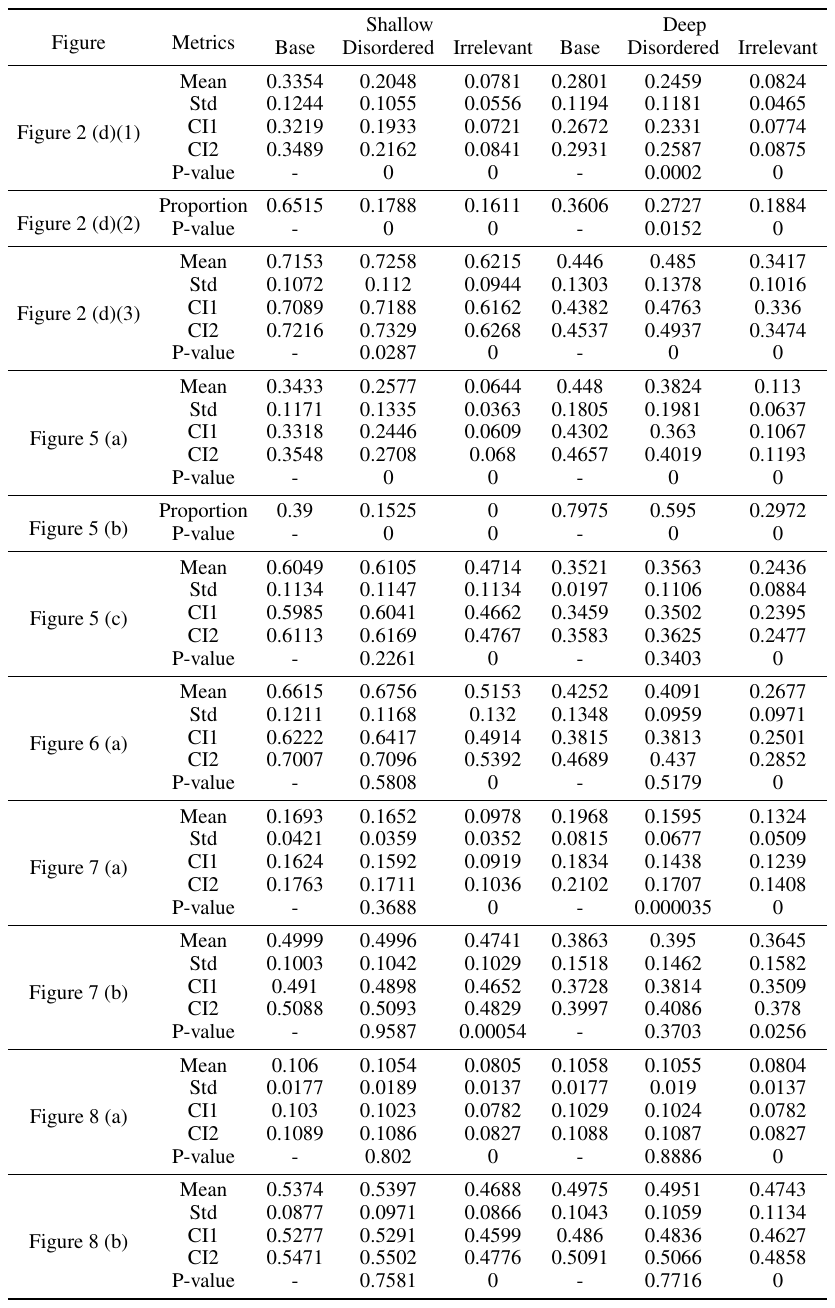}
  \vspace{-50pt}
  \caption{Saliency score analysis including mean, std, p-value, and confidence interval. Confidence interval is (CI1, CI2). P-value is calculated based on value of Base.}
\label{fig:fig_mean_analysis}
\end{figure*}

\subsubsection{Additional analysis}

%we analyze the information flow based on saliency score when LLM is confronted with inputs that are either disordered or peppered with irrelevant information from multiple models and datasets.

In this section, we analyze the information flow based on saliency score on multiple models and datasets, to further support the observations in section\ref{sal_analysis}. We perform experiments on the ProofWriter and DI-GSM dataset based on Llama-2-13B-Chat and Qwen-14B-Chat. Figure \ref{fig:fig_score_qwen} shows the saliency score analysis on ProofWriter based on Qwen-14B-Chat. Consistent with the analysis on ProofWriter based on Llama-2-13B-Chat, when inputs are concise and organized, the saliency score from the ground-truth reasoning entrance to the first reasoning step is significantly higher than when inputs are disordered or contain irrelevant information, as shown in Figure \ref{fig:fig_score_qwen}(a). The proportion of samples with the highest saliency score from the ground-truth reasoning entrance to the first reasoning step is still the highest when inputs are concise and organized. As shown in Figure \ref{fig:fig_score_qwen}(b), in the shallow layer analysis, the proportion when inputs are concise and organized is as high as 40\%, while the proportion when inputs contain irrelevant information is 0\%. In the deep layer analysis, the proportion when inputs are concise and organized is as high as 80\%, while the proportion when inputs are disordered or contain irrelevant information drops significantly. This analysis suggests that concise and organized inputs will empower the model to accurately identify the entry point for reasoning, thus minimizing failures attributed to inaccuracies at the reasoning entrance. Besides, information flow from the previous two steps to the current step and irrelevant information is also relatively salient, which is also consistent with the previous analysis. On the one hand, information flow from previous two steps to the current step is salient in Figure \ref{fig:fig_score_qwen}(c), which complicates the identification of the next correct reasoning path, and even makes up premises when faced with disordered and irrelevant content. On the other hand, as shown in Figure \ref{fig:fig_score_qwen}(d), salient information flow from irrelevant information makes models distractible, causing models to focus on irrelevant content and leading to reasoning failure.

In addition, we further analyze the generated fake premises when reasoning on ProofWriter based on Qwen-14B-Chat, which can be considered as hallucination. As shown in Figure \ref{fig:fig_gen_qwen_count}(b), the number of the generated fake premises in the disordered case is larger than that in the concise and organized case, and the number of the generated fake premises in the irrelevant case is as high as 120. Comparing Figure \ref{fig:fig_score_qwen}(c) and Figure \ref{fig:fig_gen_qwen_count}(a), information flow from the previous two steps to the current step is more salient when generating fake premises than the average. Information flow from irrelevant information in Figure \ref{fig:fig_gen_qwen_count}(c) has the same trend, with an increase of about 10\% in both shallow and deep layers. Figure \ref{fig:fig_score_llama_gs} and Figure \ref{fig:fig_score_qwen_gs} show the saliency score analysis on DI-GSM based on Llama-2-13B-Chat and Qwen-14B-Chat. Similar results occurred on these models and data, further revealing the negative impact of disordered and irrelevant information on reasoning.

\subsubsection{Other statistical analysis}
Other statistical analysis including std, p-value and confidence interval are shown in Figure ~\ref{fig:fig_mean_analysis}. Column "Figures" can be linked to figures in main text. Results show that saliency score analysis based on mean metric can directly reflect differences under different experimental settings to a certain extent.

%\clearpage

\subsection{Additional experimental analysis}
\label{sec:sec_append_exps}

\subsubsection{Performance on different parts of ProofWriter and PrOntoQA.}
Table ~\ref{tab:tab_different_part} lists the detailed results on different parts of ProofWriter and PrOntoQA. COP outperforms SOTA
methods on different reasoning depths (like easy 1-hop, hard 5-hop) on PrOntoQA. For different parts of ProofWriter, COP is still the best.
\label{sec:sec_different_part}
% Table generated by Excel2LaTeX from sheet 'Sheet11'
\begin{table*}[htbp]
  \small
  \centering
  \caption{Performance on different parts of ProofWriter and PrOntoQA.}
    \begin{tabular}{c|cccccc|cccc}
    \toprule
    Datasets/ & \multicolumn{6}{c|}{ProofWriter}              & \multicolumn{4}{c|}{PrOntoQA} \\
    Methods & d5    & d3    & d2    & d1    & d0    & average & 5-hop & 3-hop & 1-hop & average \\
    \midrule
    %Standard & 41.67  & 49.83  & 51.00  & 55.50  & 63.67  & 52.33  & 49.60  & 52.00  & 65.60  & 55.73  \\
    CoT   & 53.50  & 61.17  & 61.33  & 62.33  & 62.83  & 60.23  & 69.80  & 74.20  & 86.20  & 76.73  \\
    SI    & 46.00  & 51.00  & 56.00  & 61.00  & 97.00  & 62.20  & 45.00  & 52.00  & 97.00  & 64.67  \\
    LogicLM & 70.11  & -     & -     & -     & -     & -     & 93.20  & -     & -     & - \\
    LAMBABDA & 72.00  & 82.00  & 87.00  & 90.00  & 98.00  & 85.80  & 96.00  & 99.00  & 98.00  & 97.67  \\
    \textbf{COP} & \textbf{88.67 } & \textbf{90.67 } & \textbf{91.43 } & \textbf{92.50 } & \textbf{98.50 } & \textbf{91.72 } & \textbf{99.20 } & \textbf{99.60 } & \textbf{100.00 } & \textbf{99.60 } \\
    \bottomrule
    \end{tabular}%
  \label{tab:tab_different_part}%
\end{table*}%

\subsubsection{Performance with different large language models}
\label{sec:sec_llms}
To study if the proposed COP is effective across different base models, we test our method on several closed-source LLMs including GPT-3.5-turbo, GPT-4o, Claude-3-5-Sonnet and Gemini-1.0-Pro in main text. In addition, we conducted experiments on DI-GSM with several weaker LLMs including Qwen1.5-72B-Chat, Llama-2-13B-Chat, mistral-7b-instruct-v0.3, Gemini-1.0-Pro and GPT-4o-mini. As shown in Table~\ref{tab:tab_llms}, COP still consistently achieves high accuracy using different LLMs, revealing its effectiveness across different LLMs (open-source and close-source LLMs).

% Table generated by Excel2LaTeX from sheet 'Sheet3'
\begin{table*}[htbp]
  \centering
  \small
  \caption{The performance comparisons on DI-GSM using different weaker close-source and open-source LLMs.}
    \begin{tabular}{c|ccccc}
    \toprule
    Methods & \tabincell{c}{Qwen1.5\\-72B-Chat} & \tabincell{c}{Llama-2\\-13B-Chat} & \tabincell{c}{mistral-7b\\-instruct-v0.3} & \tabincell{c}{Gemini-1.0\\-Pro} & \tabincell{c}{GPT-4o\\-mini} \\
    \midrule
    CoT   & 56.82  & 11.36  & 20.45  & 36.36  & 70.45  \\
    \textbf{COP} & \textbf{59.09 } & \textbf{15.15 } & \textbf{27.27 } & \textbf{55.30 } & \textbf{71.97 } \\
    \bottomrule
    \end{tabular}%
  \label{tab:tab_llms}%
\end{table*}%

\iffalse
% Table generated by Excel2LaTeX from sheet 'Sheet3'
\begin{table*}[htbp]
  \centering
  \small
  \caption{The performance comparisons on DI-GSM using different LLMs.}
    \begin{tabular}{c|cccc}
    \toprule
    Methods & \tabincell{c}{Qwen2\\-72B-Instruct} & \tabincell{c}{Llama-3\\-70B-Instruct} & \tabincell{c}{Llama-3\\-8B-Instruct} & \tabincell{c}{Llama-2\\-13B-Chat} \\
    \midrule
    CoT   & 78.79 & 72.73 & 43.18 & 11.36 \\
    \textbf{COP}   & \textbf{80.30} & \textbf{78.79} & \textbf{52.27} & \textbf{15.15} \\
    \bottomrule
    \end{tabular}%
  \label{tab:tab_llms}%
\end{table*}%
\fi

% Table generated by Excel2LaTeX from sheet 'Sheet5'
\begin{table*}[htbp]
  \centering
  \caption{Comparison of average token numbers on the ProofWriter dataset.}
    \begin{tabular}{ccccccc}
    \toprule
    \textcolor[rgb]{ .2,  .2,  .2}{Hops} & \textcolor[rgb]{ .2,  .2,  .2}{0} & \textcolor[rgb]{ .2,  .2,  .2}{1} & \textcolor[rgb]{ .2,  .2,  .2}{2} & \textcolor[rgb]{ .2,  .2,  .2}{3} & \textcolor[rgb]{ .2,  .2,  .2}{4} & \textcolor[rgb]{ .2,  .2,  .2}{5} \\
    \midrule
    \textcolor[rgb]{ .2,  .2,  .2}{LAMBADA-Prompt} & \textcolor[rgb]{ .2,  .2,  .2}{567.71} & \textcolor[rgb]{ .2,  .2,  .2}{4825.98} & \textcolor[rgb]{ .2,  .2,  .2}{8154.11} & \textcolor[rgb]{ .2,  .2,  .2}{9247.04} & \textcolor[rgb]{ .2,  .2,  .2}{14401.85} & \textcolor[rgb]{ .2,  .2,  .2}{19200.05} \\
    \textcolor[rgb]{ .2,  .2,  .2}{LAMBADA-Total} & \textcolor[rgb]{ .2,  .2,  .2}{611.76} & \textcolor[rgb]{ .2,  .2,  .2}{5293.22} & \textcolor[rgb]{ .2,  .2,  .2}{8992.39} & \textcolor[rgb]{ .2,  .2,  .2}{10333.2} & \textcolor[rgb]{ .2,  .2,  .2}{15944.14} & \textcolor[rgb]{ .2,  .2,  .2}{21922.77} \\
    \textcolor[rgb]{ .2,  .2,  .2}{COP-Prompt} & \textcolor[rgb]{ .2,  .2,  .2}{433.21} & \textcolor[rgb]{ .2,  .2,  .2}{1876.82} & \textcolor[rgb]{ .2,  .2,  .2}{1915.68} & \textcolor[rgb]{ .2,  .2,  .2}{1953.45} & \textcolor[rgb]{ .2,  .2,  .2}{1996.62} & \textcolor[rgb]{ .2,  .2,  .2}{2004.97} \\
    \textcolor[rgb]{ .2,  .2,  .2}{COP-Total} & \textcolor[rgb]{ .2,  .2,  .2}{594.53} & \textcolor[rgb]{ .2,  .2,  .2}{2199.44} & \textcolor[rgb]{ .2,  .2,  .2}{2270.31} & \textcolor[rgb]{ .2,  .2,  .2}{2341.29} & \textcolor[rgb]{ .2,  .2,  .2}{2425.71} & \textcolor[rgb]{ .2,  .2,  .2}{2440.26} \\
    \bottomrule
    \end{tabular}%
  \label{tab:tab_tokens}%
\end{table*}%

% Table generated by Excel2LaTeX from sheet 'Sheet6'
\begin{table*}[htbp]
  \centering
  \caption{Comparison of average inference calls and token numbers on the FOLIO dataset.}
    \begin{tabular}{cccc}
    \toprule
    Method & Calls  & Prompt-tokens & Total-tokens \\
    \midrule
    LogicLM & 3.88     & 6204.39 & 7281.57 \\
    COP   & 3     & 3801.99 & 4104.92 \\
    \bottomrule
    \end{tabular}%
  \label{tab:tab_tokens_folio}%
\end{table*}%

\begin{figure}[t]
  %\vspace{-10pt}
  \centering
  \includegraphics[width=0.4\textwidth]{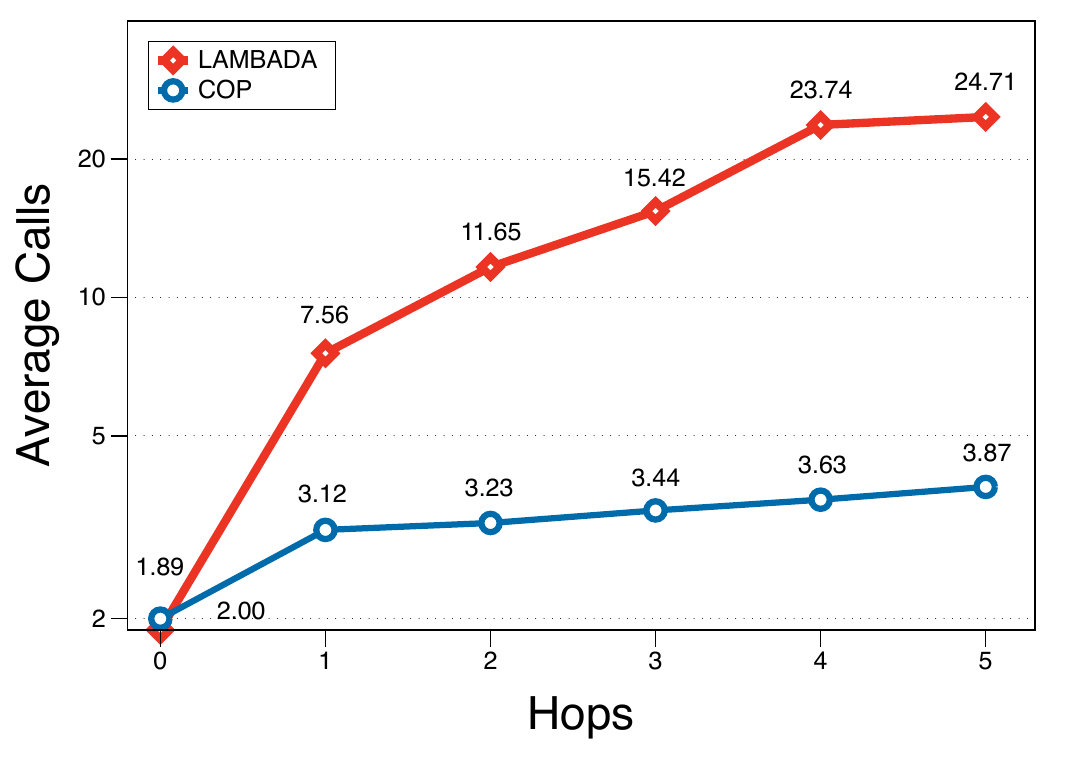}
  \caption{Comparison of inference calls on the ProofWriter dataset.}
\label{fig:fig_inference_calls}
\end{figure}

\iffalse
\begin{figure*}[ht]
    \begin{center}
    \subfigure[]{
            \includegraphics[width=0.35\linewidth]{figs/5hops.pdf}
    }%
    \subfigure[]{
            \includegraphics[width=0.3\linewidth]{figs/pie1_1.pdf}
    }%
    \subfigure[]{
            \includegraphics[width=0.3\linewidth]{figs/pie2_2.pdf}
    }%
    \end{center}
        \caption{(a) Comparison of LAMBADA and LAMBADA+COP on Proofwriter. (b) The proportions of different error reasons of LAMBADA. (c) The proportions of different error reasons of LAMBADA + COP. There are four steps in LAMBADA, which are Fact Check step, Rule Selection step, Goal Decomposition step and Sign Agreement step. "Incorrect-Selection" means that LAMBADA fails in Fact Check or Rule Selection steps. "Incorrect-Goal" means that LAMBADA fails in Goal Decomposition steps. "Incorrect-Sign" means that LAMBADA fails in Sign Agreement steps.}
        \label{fig:fig_append}
    \vspace{-10pt}
\end{figure*}
\fi

\subsubsection{Number of inference calls and tokens}
\label{sec:sec_append_tokens}
In Figure \ref{fig:fig_inference_calls}, we compared the average number of inference calls per example under different reasoning depths on the ProofWriter dataset. COP requires significantly fewer inference calls than LAMBADA, and the number of inference calls remains relatively stable as the number of hops increases. Besides, Table \ref{tab:tab_tokens} compares token numbers used per question on the ProofWriter dataset with different hops. The token numbers are taken from the usage statistics returned by the OpenAI API. COP-Prompt and LAMBADA-Prompt stand for the input token numbers of COP and LAMBADA, while COP-Total and LAMBADA-Total stand for the overall token consumed by input and output. As shown in the table, COP costs much fewer token numbers than LAMBADA, and the number of token numbers remains relatively stable as the number of hops increases, demonstrating our proposed COP’s superiority in both effectiveness and efficiency.

While LAMBADA is not able to work on DI-GSM and FOLIO, and LogicLM is not able to work on DI-GSM, we compared the average number of inference calls and tokens used in COP and LogicLM on FOLIO. Table \ref{tab:tab_tokens_folio} shows the results. COP still costs fewer inference calls and much fewer token numbers than LogicLM, which demonstrates the superiority of COP in terms of efficiency.

\subsubsection{Concise and organized perception analysis}
\label{sec:sec_realmetric}
As analyzed in Section\ref{sal_analysis}, disordered and irrelevant content has a negative impact on reasoning from the perspective of information flow. The proposed COP seeks to enhance reasoning performance and simplify the reasoning process by systematically removing irrelevant information and reorganizing the inputs. Thus, we analyze the performance of COP from two aspects: the degree of removal of irrelevant information and the order of input. Firstly, each sample in DI-GSM contains 2 to 3 irrelevant statements to the questions, for a total of 322 irrelevant statements. After applying COP, only nine statements have not been removed, demonstrating the success of concise perception in COP. Secondly, the statements of questions are randomly shuffled in DI-GSM. We use Kendall tau distance to measure the order difference of statements between the shuffled data and the original data. Kendall tau distance ranges from -1 to 1. The larger the Kendall tau distance, the more relevant it is. Before applying COP, the average Kendall tau distance is -0.0465, indicating no strong correlation between the statement order in the shuffled data and the original data. After applying COP, the average Kendall tau distance increases to 0.4268, demonstrating the success of organized perception in COP.

\subsubsection{Proof accuracy analysis}
\label{sec:sec_proof}
Table\ref{tab:tab_res} lists the label accuracy comparison, while proof accuracy is a more stringent metric. Previous studies have demonstrated that CoT predicts a correct label with incorrect reasoning chains \citep{saparov2023language}. To validate if it is the case for COP, we randomly selected 100 correctly answered samples from the Depth-5 setting of ProofWriter. We manually checked the reasoning chain produced by LLMs with COP. According to our observation, only 7 out of 100 samples contain invalid reasoning steps, which indicates that the proposed COP does arouse the reasoning ability of LLMs, and the experimental results reported above are faithful.

%\iffalse
\subsubsection{Is COP beneficial to other methods ?}
\label{sec:sec_error_other}
Based on \textit{easy to plan}, our COP can be seamlessly combined with methods that teach models \textit{how to plan}, such as LAMBADA. Since LAMBADA cannot work on two real-world datasets, FOLIO and DI-GSM, we use ProofWriter for testing. The performance of LAMBADA and LAMBADA+COP on the ProofWriter $d_5$ subset with different inference depths are listed in Figure \ref{fig:fig_append} (a). All the results are based on the LAMBADA code we reproduced, and the base model of this experiment is GPT-3.5-turbo. Compared with the original LAMBADA method, the performance of LAMBADA+COP under different inference depths is improved, proving the effectiveness of COP. In addition, Figure \ref{fig:fig_append}(b) shows the proportion of correct reasoning and the proportion of different types of incorrect reasoning. We randomly selected 100 test samples from the ProofWriter $d_5$ subset to manually check the error types of incorrect reasoning examples. As shown in the figure, equipped with COP, the proportion of selection errors (including fact check and rule selection modules in LAMBADA) drops significantly. The proportion of goal decomposition errors and sign agreement errors (goal decomposition and sign agreement modules are unaffected by context redundancy and disorder) are almost unchanged, proving that our COP can improve the success rate of other methods in the steps that are affected by context redundancy and disorder.
%\fi

\newpage
\onecolumn

\subsection{Case study}
\label{sec:sec_case_study}
Here we present three cases to show the effectiveness and limitations of our COP. 

\textbf{Case 1:} Case 1 shows a case from Proofwriter dataset, where the input contains disorder and distractibility, making it difficult for LLM to perform reasoning. After applying COP, the input is concise and organized based on the given question. Based on the input after applying COP, LLM reasons easily, proving the effectiveness of COP.

\textbf{Case 2:} Case 2 shows a good case from DI-GSM dataset. Similar to case in Case 1, after applying COP, the input is concise and organized based on the given question. The order of statements after applying COP is different from the original order, and the changed order generated by COP is more suitable for reasoning, demonstrating the success of organized perception in COP. This is a good case mentioned in Section ~\ref{sec:sec_error_self}.

\textbf{Case 3:} As mentioned in Section ~\ref{sec:sec_error_self} Failure Case Analysis, there are 13 cases where the premises are not connected to other premises in the given context. Case 3 shows one of 13 cases. The key information "It takes 75 large jelly beans to fill Grandpa up" is not connected to other premises in COP, resulting in missing of this information in the final input after applying COP and further leading to reasoning failure. We also discuss this issue in Section Limitations, performing robust capturing of locally-related premises and generating more appropriate tree-like mind map structures require further exploration for more general reasoning tasks. We plan to address these issues in future research.

\begin{tcolorbox}[width=1.\textwidth, fontupper=\small, title = {Case 1: Input before and after applying our COP in Proofwriter}, colback=white, colbacktitle=black!5!white, coltitle=black, boxrule=1pt, ]
\textbf{Input with disorder and distractibility:} Context: If something eats the tiger then it eats the bear. The bear is young. The bear chases the dog. The bear eats the tiger. If something eats the dog then the dog is young. The bald eagle is green. The tiger chases the bear. The bear eats the bald eagle. The dog is young. If something is red and it eats the dog then the dog eats the tiger. The bear is big. The bald eagle eats the tiger. If something is rough then it eats the bear. If something visits the tiger then the tiger eats the bear. The tiger chases the dog. The bear is green. The bear chases the bald eagle. The bear eats the dog. The dog is big. If something is green and it visits the bald eagle then it visits the dog. If something eats the bear then it is red. Question: Based on the above information, is the following statement true, false, or unknown? The bald eagle does not eat the bear.

\textbf{Input after Applying COP:} Context: The bald eagle eats the tiger. If something eats the tiger then it eats the bear. Question: Based on the above information, is the following statement true, false, or unknown? The bald eagle does not eat the bear.

\textbf{Reasoning after Applying COP:} The bald eagle eats the tiger. If something eats the tiger then it eats the bear. Therefore, the bald eagle eats the bear. Therefore, the given statement 'The bald eagle does not eat the bear' is False.
\end{tcolorbox}

\begin{tcolorbox}[width=1.\textwidth, fontupper=\small, title = {Case 2: Good Case in DI-GSM}, colback=white, colbacktitle=black!5!white, coltitle=black, boxrule=1pt, ]
\textbf{Original Input:} Annabelle is saving for a phone that costs \$400. She already has \$80 in her savings. Her first job, where she earns \$10 per hour, pays her for 20 hours of work. She is also paid for 15 hours of work at her second job, where she earns \$5 an hour. In dollars, how much money does Annabelle still need to save?

\textbf{Input with disorder and distractibility:} Annabelle's first job, where Annabelle earns \$10 per hour, pays her for 20 hours of work. Annabelle already has \$80 in her savings. Annabelle is also paid for 15 hours of work at her second job, where Annabelle earns \$5 an hour. Fern is checking IDs to get into an R-rated movie. Grandpa Lou enjoys watching movies on the Hallmark channel, where every movie lasts 90 minutes. Annabelle is saving for a phone that costs \$400. In dollars, how much money does Annabelle still need to save?

\textbf{Input after Applying COP:} Annabelle's first job, where Annabelle earns \$10 per hour, pays her for 20 hours of work. Annabelle already has \$80 in her savings. Annabelle is also paid for 15 hours of work at her second job, where Annabelle earns \$5 an hour. Annabelle is saving for a phone that costs \$400. In dollars, how much money does Annabelle still need to save?
\end{tcolorbox}

\begin{tcolorbox}[width=1.\textwidth, fontupper=\small, title = {Case 3: Bad Case in DI-GSM}, colback=white, colbacktitle=black!5!white, coltitle=black, boxrule=1pt, ]
\textbf{Original Input:} Grandpa loves to eat jelly beans, but how many jelly beans he can eat depends on the size of the beans. \underline{It takes 75 large jelly beans to fill Grandpa up.} He can eat twice as many medium-sized beans as large beans. And eating 3 small beans is the same as eating 1 medium-sized bean. How many small beans can Grandpa eat?

\textbf{Input with disorder and distractibility:} There are 5 houses on a street, and each of the first four houses has 3 gnomes in the garden. Grandpa can eat twice as many medium-sized beans as large beans. It takes 75 large jelly beans to fill Grandpa up. Last year Jessica paid \$1000 for rent, \$200 for food, and \$100 for car insurance each month. And eating 3 small be ans is the same as eating 1 medium-sized bean. Hans booked a room in a hotel. Grandpa loves to eat jelly beans, but how many jelly beans Grandpa can eat depends on the size of the beans. How many small beans can Grandpa eat?

\textbf{Input after Applying COP:} Grandpa loves to eat jelly beans, but how many jelly beans Grandpa can eat depends on the size of the beans. Grandpa can eat twice as many medium-sized beans as large beans. Eating 3 small beans is the same as eating 1 medium-sized bean. How many small beans can Grandpa eat?
\end{tcolorbox}

\subsection{Prompts used in experiments}
\label{sec:sec_append_prompts}
In this section, we give the prompts used in our experiments. The prompts used in PrOntoQA and PrOntoQA-OOD are similar to that in ProofWriter.
%\begin{tcolorbox}[title = {Prompts for GSM8K}, colback=white, colbacktitle=black!5!white, coltitle=black, boxrule=1pt, ]

%\end{tcolorbox}

%\newpage

\subsubsection{Prompts in capturing of locally-related premises}
\begin{tcolorbox}[width=1.\textwidth, fontupper=\small, title = {Prompts for DI-GSM [Capturing of locally-related premises]}, colback=white, colbacktitle=black!5!white, coltitle=black, boxrule=1pt, ]
Given multiple statements in a context, the task is to find relevant statements for each statement. Use "A -> B" to denote statement B that is relevant after statement A. Use "A -> None" to denote that there is no statement that is relevant after statement A. Each statement can have multiple relevant statements. Do not change the logic and content of the statements in context.\\
------\\
Context:\\
James makes potatoes for a group.\\
For every 5 fruits that customers buy, the store offers a \$1 discount.\\
Mary went to the store to buy fruit.\\
Each person eats 1.5 pounds of potatoes.\\
Apples cost \$1, oranges cost \$2, and bananas cost \$3.\\
Mary buys 5 apples, 3 oranges, and 2 bananas.\\
Margaret wants to serve chicken salad sandwiches using mini croissants.\\

Answer:\\
James makes potatoes for a group. -> Each person eats 1.5 pounds of potatoes.\\
For every 5 fruits that customers buy, the store offers a \$1 discount. -> None.\\
Mary went to the store to buy fruit. -> Mary buys 5 apples, 3 oranges, and 2 bananas.\\
Each person eats 1.5 pounds of potatoes. -> None.\\
Apples cost \$1, oranges cost \$2, and bananas cost \$3. -> For every 5 fruits that customers buy, the store offers a \$1 discount.\\
Mary buys 5 apples, 3 oranges, and 2 bananas. -> Apples cost \$1, oranges cost \$2, and bananas cost \$3.\\
Margaret wants to serve chicken salad sandwiches using mini croissants. -> None.\\

------\\
Context:\\
...\\

------\\
Context:\\
The middle height tree is 2/3 the height of the tallest tree.\\
At the burger hut, you can buy a burger for \$5, french fries for \$3, and a soft drink for \$3.\\
There are three trees in the town square.\\
The tallest tree is 150 feet tall.\\
George is about to celebrate his 25th birthday.\\
The shortest tree is half the size of the middle tree.\\

Answer:\\
The middle height tree is 2/3 the height of the tallest tree. -> The shortest tree is half the size of the middle tree.\\
At the burger hut, you can buy a burger for \$5, french fries for \$3, and a soft drink for \$3. -> None.\\
There are three trees in the town square. -> The tallest tree is 150 feet tall.\\
The tallest tree is 150 feet tall. -> The middle height tree is 2/3 the height of the tallest tree. \\
George is about to celebrate his 25th birthday. -> None.\\
The shortest tree is half the size of the middle tree. -> None.\\

------\\
Context:

[[PREMISES]]\\

Answer:
\end{tcolorbox}

%\onecolumn
%\newpage

\begin{tcolorbox}[width=1.\textwidth, fontupper=\small, title = {Prompts for FOLIO [Capturing of locally-related premises]}, colback=white, colbacktitle=black!5!white, coltitle=black, boxrule=1pt, ]
Given multiple statements in a context, the task is to find logically relevant statements for each statement. Use "A -> B" to denote statement B that is logically relevant after statement A. Use "A -> None" to denote that there is no statement that is logically relevant after statement A. Each statement can have multiple logically relevant statements. Do not change the logic and content of the statements in context.\\
------\\
Context:\\
A thing is either a plant or animal. \\
If a sea eel is either an eel or a plant, then a sea eel is an eel or an animal.\\
No fish are plants. \\
All animals breathe. \\
Nothing that breathes is paper. \\
All eels are fish. \\

Answer:\\
A thing is either a plant or animal. -> All animals breathe.\\
If a sea eel is either an eel or a plant, then a sea eel is an eel or an animal. -> All eels are fish.\\
If a sea eel is either an eel or a plant, then a sea eel is an eel or an animal. -> All animals breathe.\\
No fish are plants. -> None. \\
All animals breathe. -> Nothing that breathes is paper.\\
Nothing that breathes is paper. -> None.\\
All eels are fish. -> No fish are plants.\\

------\\
Context:\\
...\\

------\\
Context:\\
All Instagram is entertainment. \\
All video applications are software. \\
If something is interesting, then it is good. \\
All YouTube-related applications are video applications. \\
All entertainments are interesting. \\
TikTok is not good.\\
All software is programmed. \\
An APP is either related to YouTube or Instagram. \\

Answer:\\
All Instagram is entertainment. -> All entertainments are interesting. \\
All video applications are software. -> All software is programmed.\\
If something is interesting, then it is good. -> TikTok is not good.\\
All YouTube-related applications are video applications. -> All video applications are software.\\
All entertainments are interesting. -> If something is interesting, then it is good.\\
TikTok is not good. -> None. \\
All software is programmed. -> None.\\
An APP is either related to YouTube or Instagram. -> All YouTube-related applications are video applications.\\
An APP is either related to YouTube or Instagram. -> All Instagram is entertainment.\\

------\\
Context:

[[PREMISES]]\\

Answer:
\end{tcolorbox}

\begin{tcolorbox}[width=1.\textwidth, fontupper=\small, title = {Prompts for Proofwriter [Capturing of locally-related premises / Premises Unified Formats]}, colback=white, colbacktitle=black!5!white, coltitle=black, boxrule=1pt, ]
You are given some known rules. Extract the conditions and consequents of each rule and output follow the format of the given examples:\\
Examples:\\
Rules:\\
If someone sees the cat and they are not green then they see the cow. If the rabbit is kind and the rabbit sees the squirrel then the squirrel needs the rabbit. Rough people are cold. If someone sees the rabbit then they are not round. If someone sees the squirrel and they are not green then they need the squirrel. If someone eats the cow then they see the rabbit. Cold things are rough. If someone is cold then they eat the cow. Kind, rough people are round.\\
Output:\\
{”Rule1”: {”conditions”: [”X(see, cat)”, ”X(is not, green)”], ”consequents”: [”X(see, cow)”]}, ”Rule2”: {”conditions”: [”rabbit(is, kind)”, ”rabbit(see, squirrel)”], ”consequents”: [”squirrel(need, rabbit)”]}, [...], ”Rule9”: {”conditions”: [”X(is, kind)”, ”X(is, rough)”], ”consequent”: [”X(is, round)”]}}\\

Rules:\\
...\\

Rules:\\
If something visits the mouse and the mouse visits the dog then it is cold. If mouse likes the cat then it visits the dog. If something is cold then it likes the cat. If something is green then it sees the dog. If something likes the mouse then it sees the cat. If dog is green and cold then it likes the cat. If something is big and it visits the bear then the bear is green. Round things are rough.
Output: \\
{”Rule1”: {”conditions”: [”X(visit, mouse)”, ”mouse(visit, dog)”], ”consequents”: [”X(is, cold)”]}, ”Rule2”: {”conditions”: [”mouse(like, cat)”], ”consequents”: [”X(visit, dog)”]}, ”Rule8”: {”conditions”: [”X(is, round)”], ”consequents”: [”X(is, rough)”]}}\\

Rules:

[[PREMISES]]\\

Output:
\end{tcolorbox}

\begin{tcolorbox}[width=1.\textwidth, fontupper=\small, title = {Prompts for Proofwriter [Capturing of locally-related premises / Premises Unified Formats]}, colback=white, colbacktitle=black!5!white, coltitle=black, boxrule=1pt, ]
You are given some known facts. Output the facts following the format of the given examples:\\
Examples:\\
Facts:\\
The bear is green. The bear likes the cat. The bear likes the dog. The bear visits the dog. The cat isyoung. The cat does not see the bear. The cat sees the dog. The cat visits the bear. The dog is round. The mouse is not big. The mouse is cold.\\
Output: \\
”Fact1”: [”bear(is, green)”], ”Fact2”: [”bear(like, cat)”], ”Fact3”: [”bear(like, dog)”], ”Fact4”: [”bear(visit, dog)”], ”Fact5”: [”cat(is, young)”], ”Fact6”: [”cat(not see, bear)”], ”Fact7”: [”cat(see, dog)”], ”Fact8”: [”cat(visit, bear)”], ”Fact9”: [”dog(is, round)”], ”Fact10”: [”mouse(is not, big”], ”Fact11”: [”mouse(is, cold)”]\\

Facts:\\
...\\

Facts: 
[[PREMISES]]\\

Output: 
\end{tcolorbox}

\subsubsection{Prompts in generation of mind map}
\begin{tcolorbox}[width=1.\textwidth, fontupper=\small, title = {Prompts for DI-GSM [Generation of Mind Map]}, colback=white, colbacktitle=black!5!white, coltitle=black, boxrule=1pt, ]
Given a question and multiple ordered relevant sentences in context, the task is to find in order all sentences in the context that are required to answer the given question or are related to the information and subject in the given question.\\
------\\
Context:\\
Mary went to the store to buy fruit. Mary buys 5 apples, 3 oranges, and 2 bananas. Apples cost \$1, oranges cost \$2, and bananas cost \$3.\\
James makes potatoes for a group. Each person eats 1.5 pounds of potatoes.\\
For every 5 fruits that customers buy, the store offers a \$1 discount.\\
Margaret wants to serve chicken salad sandwiches using mini croissants.\\

Question:\\
How much will Mary pay?\\

Inference:\\
All sentences in order that are required to answer the given question or are related to the information and subject in the given question are:
Mary went to the store to buy fruit. -> Mary buys 5 apples, 3 oranges, and 2 bananas. -> Apples cost \$1, oranges cost \$2, and bananas cost \$3. -> For every 5 fruits that customers buy, the store offers a \$1 discount.\\

------\\
Context:\\
...\\

------\\
Context:\\
At the burger hut, you can buy a burger for \$5, french fries for \$3, and a soft drink for \$3.\\
The tallest tree is 150 feet tall. The middle height tree is 2/3 the height of the tallest tree. The shortest tree is half the size of the middle tree.\\
George is about to celebrate his 25th birthday.\\
There are three trees in the town square. \\

Question:\\
How tall is the shortest tree?\\

Inference:\\
All sentences in order that are required to answer the given question or are related to the information and subject in the given question are:
There are three trees in the town square. -> The tallest tree is 150 feet tall. -> The middle height tree is 2/3 the height of the tallest tree. -> The shortest tree is half the size of the middle tree.\\

------\\
Context:

[[PREMISES]]\\

Question:

[[QUESTION]]\\

Inference:
\end{tcolorbox}

\begin{tcolorbox}[width=1.\textwidth, fontupper=\small, title = {Prompts for FOLIO [Generation of Mind Map]}, colback=white, colbacktitle=black!5!white, coltitle=black, boxrule=1pt, ]
Given multiple ordered logical paths in context and a statement, the task is to find the most logically relevant paths for the statement and remove irrelevant logical content in context for the statement.\\
------\\
Context:\\
If a sea eel is either an eel or a plant, then a sea eel is an eel or an animal. All eels are fish. No fish are plants.\\
If a sea eel is either an eel or a plant, then a sea eel is an eel or an animal. All animals breathe.\\
A thing is either a plant or animal. All animals breathe. Nothing that breathes is paper.\\

Statement:\\
Sea eel is a paper.\\

Answer:\\
A thing is either a plant or animal. All animals breathe. Nothing that breathes is paper. If a sea eel is either an eel or a plant, then a sea eel is an eel or an animal. All animals breathe. All eels are fish. No fish are plants.\\

------\\
Context:\\
...\\

------\\
Context:\\
An APP is either related to YouTube or Instagram. All YouTube-related applications are video applications. All video applications are software. All software is programmed.\\
An APP is either related to YouTube or Instagram. All Instagram is entertainment. All entertainments are interesting. If something is interesting, then it is good. TikTok is not good.\\

Statement:\\
Tiktok is a program.\\

Answer:\\
An APP is either related to YouTube or Instagram. All Instagram is entertainment. All entertainments are interesting. If something is interesting, then it is good. TikTok is not good. All YouTube-related applications are video applications. All video applications are software. All software is programmed.\\

------\\
Context:

[[PREMISES]]\\

Statement

[[QUESTION]]\\

Answer:
\end{tcolorbox}

\begin{tcolorbox}[width=1.\textwidth, fontupper=\small, title = {Prompts for ProofWriter [Generation of Mind Map / Question Unified Formats]}, colback=white, colbacktitle=black!5!white, coltitle=black, boxrule=1pt, ]
You are given some known facts. Output the facts following the format of the given examples:\\
Examples:\\
Facts:\\
The bear is green.\\
Output:

[”bear(is, green)”, ”bear(is not, green)”]\\

Facts:\\
...\\

Facts:\\
The mouse is not big.
Output:

[”mouse(is not, big)”, ”mouse(is, big)”]\\
Facts:

[[QUESTION]]\\

Output:
\end{tcolorbox}

\subsubsection{Prompts in CoT}
% #################### GSM8K ####################
\begin{tcolorbox}[width=1.\textwidth, fontupper=\small, title = {Prompts for DI-GSM [CoT Reasoning]}, colback=white, colbacktitle=black!5!white, coltitle=black, boxrule=1pt, ]
Question: Karen's students are about to take a standardized test. Karen gets a \$500 bonus if their average score is above 75, plus an extra \$10 bonus for every additional point the average score increases above 75. So far, Karen has graded 8 tests, and the average is 70. Given that each student can have a maximum score of 150, what combined score do the last two tests need to have for Karen to earn a \$600 bonus?
\\

Answer: First subtract \$500 from Karen's goal bonus amount to find how much she makes from the extra \$10/point bonus: \$600 - \$500 = \$<<600-500=100>>100$\backslash$nThen divide the extra bonus by the extra rate: \$100 / \$10/point = <<100/10=10>>10 points$\backslash$nThen add the 10 extra points to the baseline 75 point goal to find the students' average test score: 10 points + 75 points = <<10+75=85>>85 points$\backslash$nThen added the 8 graded tests to the 2 ungraded tests to find the total number of tests: 2 tests + 8 tests = <<2+8=10>>10 tests$\backslash$nThen multiply the 85 point average by the number of tests to find the total number of points the students need to earn: 85 points/test * 10 tests = 850 points$\backslash$nThen multiply the current average by the current number of graded tests to find how many points have been earned so far: 70 points/test * 8 tests = <<70*8=560>>560 points$\backslash$nThen subtract the number of points earned from the number of points needed to find the combine score the last two tests need: 850 points - 560 points = <<850-560=290>>290 points. The answer is 290.
\\

Question: Mr. Manuel is a campsite manager who's been tasked with checking the number of tents set up in the recreation area. On a particular day, he counted 100 tents in the northernmost part of the campsite and twice that number on the east side of the grounds. The number of tents at the center of the camp was four times the number of tents in the northernmost part of the campsite. If he also counted 200 tents in the southern part of the campsite, what is the total number of tents in the recreation area?
\\

Answer: On the eastern part of the campsite, Mr Manuel counted 2*100 = <<2*100=200>>200 tents$\backslash$nThe total number of tents in the eastern and the northern part of the campgrounds is 200+100 = <<200+100=300>>300$\backslash$nThere are four times as many tents as the northernmost part in the central part of the campsite, which means there are 4*100 = <<4*100=400>>400 tents in the central part of the camp.$\backslash$nThe total number of tents in the three parts of the campsite is 400+300 = <<400+300=700>>700$\backslash$nIf you add the number of tents Mr. Manuel counted at the southern part of the campsite, you get 700+200 = <<700+200=900>>900 tents on the whole campsite. The answer is 900.
\\

Question: [[PREMISES]][[QUESTION]]
\\

Answer:
\end{tcolorbox}

% #################### FOLIO ####################
\begin{tcolorbox}[width=1.\textwidth, fontupper=\small, title = {Prompts for FOLIO [CoT Reasoning]}, colback=white, colbacktitle=black!5!white, coltitle=black, boxrule=1pt, ]
Given a problem statement as contexts, the task is to answer a logical reasoning question. \\
------\\
Context:
The Blake McFall Company Building is a commercial warehouse listed on the National Register of Historic Places. The Blake McFall Company Building was added to the National Register of Historic Places in 1990. The Emmet Building is a five-story building in Portland, Oregon. The Emmet Building was built in 1915. The Emmet Building is another name for the Blake McFall Company Building. John works at the Emmet Building.\\

Question: Based on the above information, is the following statement true, false, or uncertain? The Blake McFall Company Building is located in Portland, Oregon.\\

Options:
A) True
B) False
C) Uncertain\\

Reasoning:
The Blake McFall Company Building is another name for the Emmet Building. The Emmet Building is located in Portland, Oregon. Therefore, the Blake McFall Company Building is located in Portland, Oregon.\\

The correct option is: A\\
------\\
Context:
People eat meat regularly or are vegetation. If people eat meat regularly, then they enjoy eating hamburgers and steaks. All people who are vegetarian are conscious of the environment or their health. If people are conscious about the environment or their health, then they do not go to fast food places often. If people have busy schedules without time to cook, then they go to fast food places often. If Jeremy does not both go to fast food places often and is conscious about the environment or their health, then he goes to fast food places often.\\

Question: Based on the above information, is the following statement true, false, or uncertain? If Jeremy has a busy schedule without time to cook, then Jeremy does not enjoy eating hamburgers and steaks.\\

Options:
A) True
B) False
C) Uncertain\\

Reasoning:
If Jeremy has a busy schedule without time to cook or enjoy eating hamburgers and steaks, then Jeremy goes to fast food places often. If people are conscious about the environment or their health, then they do not go to fast food places often. This means that Jeremy is not conscious about the environment or his health. All people who are vegetarian are conscious of the environment or their health. Therefore, Jeremy is not vegetarian. People eat meat regularly or are vegetation. Therefore, Jeremy eats meat regularly. If people eat meat regularly, then they enjoy eating hamburgers and steaks. Therefore, Jeremy enjoys eating hamburgers and steaks. \\

The correct option is: B\\
------\\
Context:
[[PREMISES]]\\

Question: [[QUESTION]]\\

Options:
A) True
B) False
C) Uncertain\\

Reasoning:
\end{tcolorbox}

% #################### ProofWriter ####################
\begin{tcolorbox}[width=1.\textwidth, fontupper=\small, title = {Prompts for ProofWriter [CoT Reasoning]}, colback=white, colbacktitle=black!5!white, coltitle=black, boxrule=1pt, ]
Given a problem statement as contexts, the task is to answer a logical reasoning question. \\
------\\
Context:
The cow is blue. The cow is round. The cow likes the lion. The cow visits the tiger. The lion is cold. The lion is nice. The lion likes the squirrel. The squirrel is round. The squirrel sees the lion. The squirrel visits the cow. The tiger likes the cow. The tiger likes the squirrel. If something is cold then it visits the tiger. If something visits the tiger then it is nice. If something sees the tiger and it is young then it is blue. If something is nice then it sees the tiger. If something likes the squirrel and it likes the cow then it visits the tiger. If something is nice and it sees the tiger then it is young. If the cow is cold and the cow visits the lion then the lion sees the squirrel.\\

Question: Based on the above information, is the following statement true, false, or unknown? The tiger is not young.\\

Options:
A) True
B) False
C) Unknown\\

Reasoning:
The tiger likes the cow. The tiger likes the squirrel. If something likes the squirrel and it likes the cow, then it visits the tiger. So the tiger visits the tiger. If something visits the tiger then it is nice. So the tiger is nice. If something is nice and it sees the tiger then it is young. So the tiger is young.\\

The correct option is: B\\
------\\
Context:
The dog sees the rabbit. The dog sees the squirrel. The dog sees the tiger. The rabbit eats the dog. The rabbit does not eat the tiger. The rabbit does not like the tiger. The squirrel does not see the rabbit. The tiger does not eat the rabbit. The tiger is not kind. The tiger likes the dog. The tiger sees the dog. If something is cold then it likes the rabbit. If something eats the tiger and it is nice then it likes the rabbit. If something likes the squirrel then the squirrel likes the rabbit. If something likes the rabbit and the rabbit is kind then it sees the tiger. If something likes the tiger then the tiger is young. If something is young and it eats the rabbit then it likes the tiger. If something sees the rabbit then the rabbit is cold. If something likes the rabbit then it likes the squirrel. If something likes the squirrel then the squirrel is cold.\\

Question: Based on the above information, is the following statement true, false, or unknown? The rabbit is cold.\\

Options:
A) True
B) False
C) Uncertain\\

Reasoning:
The dog sees the rabbit. If something sees the rabbit then the rabbit is cold. So the rabbit is cold.\\

The correct option is: A\\
------\\
Context:
[[PREMISES]]\\

Question: [[QUESTION]]\\

Options:
A) True
B) False
C) Uncertain\\

Reasoning:
\end{tcolorbox}

\end{document}